\documentclass{article}

\usepackage[final,nonatbib]{neurips_2024}

\usepackage[utf8]{inputenc} % allow utf-8 input
\usepackage[T1]{fontenc}    % use 8-bit T1 fonts
\usepackage{hyperref}       % hyperlinks
\usepackage{hyperref}
\hypersetup{
    colorlinks = true,
    urlcolor = magenta,
    citecolor = cyan
}
\usepackage{url}            % simple URL typesetting
\usepackage{booktabs}       % professional-quality tables
\usepackage{amsfonts}       % blackboard math symbols
\usepackage{nicefrac}       % compact symbols for 1/2, etc.
\usepackage{microtype}      % microtypography
\usepackage{xcolor}         % colors
\usepackage{multirow}
\usepackage{graphics}
\usepackage{amsmath}
\usepackage{graphicx}
\usepackage{algorithmic}
\usepackage{algorithm}
\usepackage{colortbl}
\usepackage{wrapfig}
\usepackage{enumitem}

\definecolor{lightgreen}{RGB}{168,209,142}
\definecolor{golden}{RGB}{255,211,71}

\title{Hamba: Single-view 3D Hand Reconstruction with Graph-guided Bi-Scanning Mamba}

\author{Haoye~Dong$^{\ast, \dagger}$ \ \  Aviral~Chharia$^{\ast}$ \ \ Wenbo~Gou$^{\ast}$ \\ \textbf{Francisco~Vicente~Carrasco \ \ \ Fernando~De~la~Torre} \\
  Carnegie Mellon University\\
  \texttt{\{haoyed, achharia, wgou, fvicente, ftorre\}@andrew.cmu.edu}\\
  \texttt{\url{https://humansensinglab.github.io/Hamba/}}
}

\begin{document}
\maketitle

\begingroup
\renewcommand\thefootnote{$*$}
\footnotetext[1]{Equal contribution. $^\dagger$Corresponding author.}

\endgroup

\begin{abstract}
3D Hand reconstruction from a single RGB image is challenging due to the articulated motion, self-occlusion, and interaction with objects. Existing SOTA methods employ attention-based transformers to learn the 3D hand pose and shape, yet they do not fully achieve robust and accurate performance, primarily due to inefficiently modeling spatial relations between joints. To address this problem, we propose a novel graph-guided Mamba framework, named \textbf{Hamba}, which bridges graph learning and state space modeling. Our core idea is to reformulate Mamba's scanning into graph-guided bidirectional scanning for 3D reconstruction using a few effective tokens. This enables us to efficiently learn the spatial relationships between joints for improving reconstruction performance. Specifically, we design a Graph-guided State Space (GSS) block that learns the graph-structured relations and spatial sequences of joints and uses 88.5\% fewer tokens than attention-based methods. Additionally, we integrate the state space features and the global features using a fusion module. By utilizing the GSS block and the fusion module, Hamba effectively leverages the graph-guided state space features and jointly considers global and local features to improve performance. Experiments on several benchmarks and in-the-wild tests demonstrate that Hamba significantly outperforms existing SOTAs, achieving the PA-MPVPE of 5.3mm and F@15mm of 0.992 on FreiHAND. At the time of this paper's acceptance, Hamba holds the top position, \textbf{Rank 1}, in two competition leaderboards\footnote{HO3Dv2 and HO3Dv3 Leaderboards: \url{https://codalab.lisn.upsaclay.fr/competitions/4318\#results}, \url{https://codalab.lisn.upsaclay.fr/competitions/4393\#results}} on 3D hand reconstruction.
\end{abstract}
%%%%%%%%%%%%%%%%%%%%%%%%%%%%%%%%%%%%%%%%%%%%%%%%%%%%%%%%%%%%%%%%%%%%%%%%%%%%%%%%%%%%%%%%%%%%%%%%%%%%%%%%%%%%%%%%%%%%%%%%%%%%%%%%%%%%%%%%%%%%%%%%%%%%%%%%%

\section{Introduction}

3D Hand reconstruction has numerous applications across multiple fields, which include robotics, animation, human-computer interaction, and AR/VR~\cite{chen2023hand,huo20233d,pei2022hand,dong2024physical,zhao2021m3d}. However, reconstructing 3D hands from a single RGB image without body context or camera parameters remains a difficult challenge in computer vision. Recent works primarily explored transformers~\cite{cho2022FastMETRO,dosovitskiy2020ViT,lin2021end,lin2021mesh, pavlakos2024reconstructing,xu2022vitpose,qian2023xformer,li2023coordinate,liu2023human} for this task and achieved SOTA performance by utilizing attention mechanism. METRO~\cite{lin2021end} introduced a multi-layer transformer, using self-attention to learn vertex-vertex and vertex-joint relations.
MeshGraphormer~\cite{lin2021mesh} integrated graph convolutions with a transformer to further enhance the reconstruction performance. Recently, HaMeR~\cite{pavlakos2024reconstructing} designed a ViT-based model~\cite{dosovitskiy2020ViT}, using ViTPose~\cite{xu2022vitpose} weights and large datasets to achieve better performance.

However, the above models fail to reconstruct a robust mesh in challenging in-the-wild scenarios that have occlusions, truncation, and hand-hand or hand-object interactions (See Figure~\ref{fig:compare} for visual comparison). This is partially due to a lack of accurate modeling of spatial relations among hand joints. Secondly, transformer-based methods~\cite{dosovitskiy2020ViT,lin2021end,lin2021mesh,pavlakos2024reconstructing,xu2022vitpose} require a large number of tokens for reconstruction, and applying attention to all image tokens does not efficiently model the joint spatial sequences (i.e., the spatial relationship between joints), which often results in an inaccurate 3D hand mesh in real-world scenarios.

\begin{figure}[t]
  \centering
  \includegraphics[width=\textwidth]{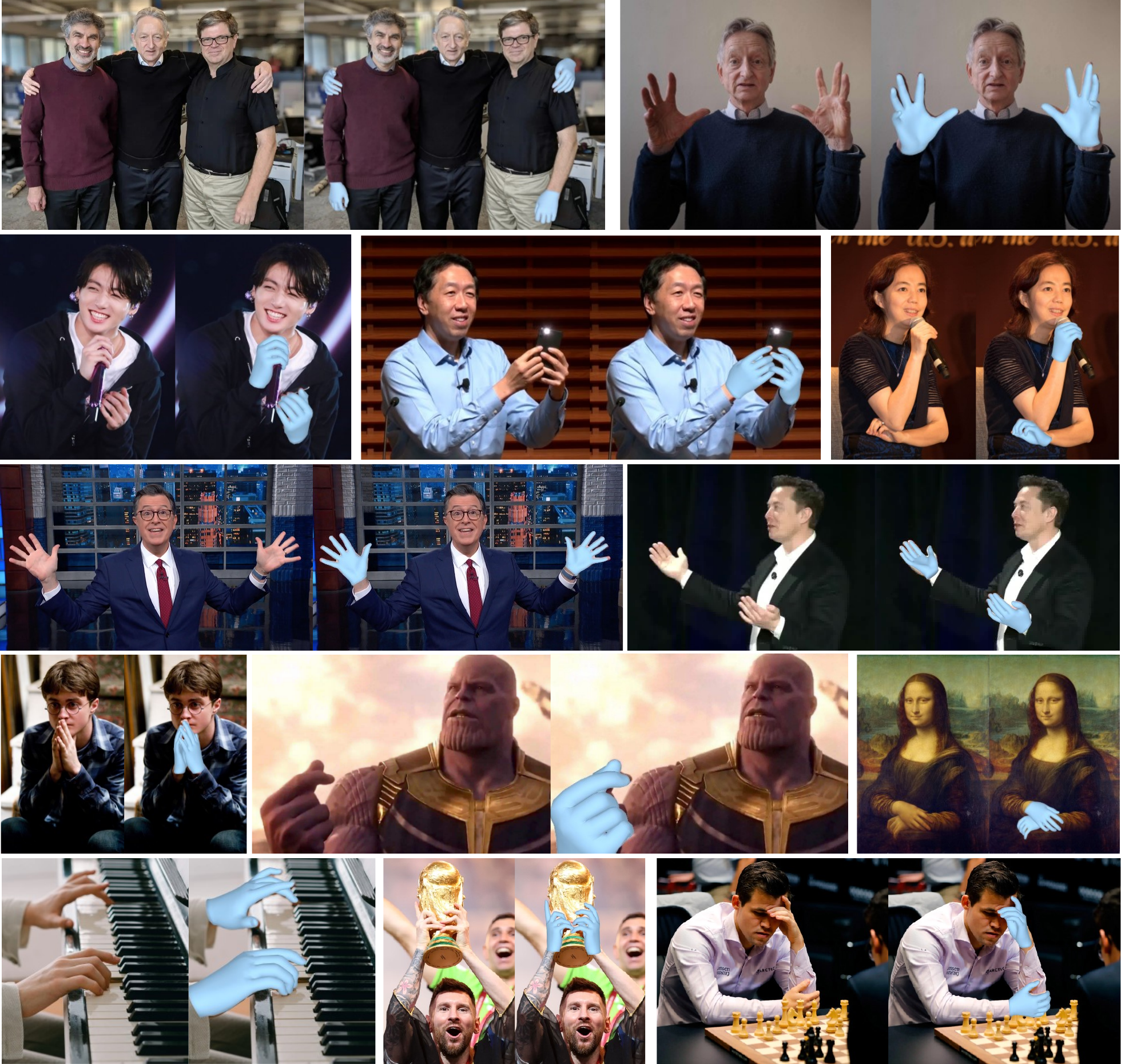}
  \vspace{-5mm}
  \caption{\textbf{In-the-wild visual results of Hamba}. Hamba achieves significant performance in various in-the-wild scenarios, including hand interaction with objects or hands, different skin tones, different angles, challenging paintings, and vivid animations.}
  \label{fig:teaser}
  \vspace{-6mm}
\end{figure}

\begin{figure}[t]
  \centering
  \includegraphics[width=0.90\textwidth]{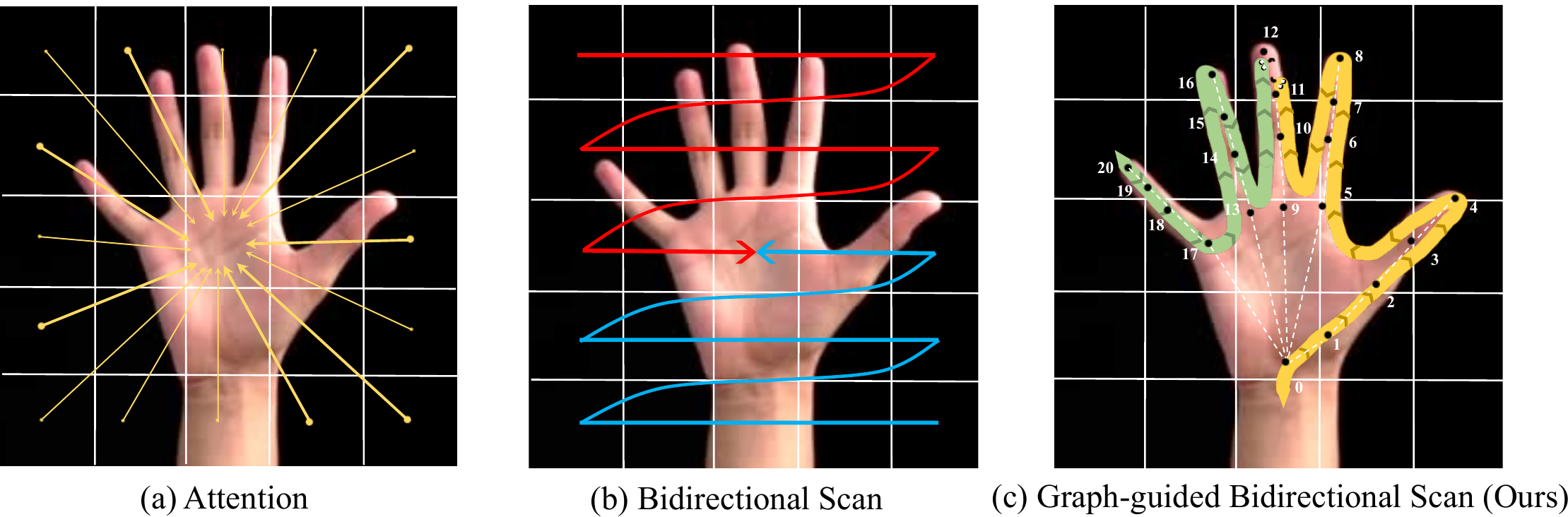}
  \vspace{-2mm}
  \caption{\textbf{Motivation}. Visual comparisons of different scanning flows. (a) Attention methods compute the correlation across all patches leading to a very high number of tokens. (b) Bidirectional scans follow two paths, resulting in less complexity. (c) The proposed graph-guided bidirectional scan (GBS) achieves effective state space modeling leveraging graph learning with a few effective tokens (illustrated as scanning by two snakes: \textcolor{golden}{forward} and \textcolor{lightgreen}{backward} scanning snakes).}
  \label{fig:motivation}
  \vspace{-3mm}
\end{figure}

To address these challenges, we propose \textbf{Hamba}, a novel Mamba-based~\cite{gu2023mamba} model that employs graph learning~\cite{lee2018higherorder,zhaoCVPR19semantic} and state space modeling~\cite{gu2023mamba} for robust 3D hand mesh reconstruction. Mamba is a new state space modeling method with global receptive field capability. Most Mamba-based models~\cite{behrouz2024graph,gu2023mamba,hu2024zigma,lieber2024jamba,wang2024graphmamba,wang2024graph,zhang2024motion} are designed for long-range data, and few studies~\cite{liang2024pointmamba,shen2024gamba} have adapted Mamba for 3D vision tasks. In this work, we explore Mamba's potential for the 3D hand reconstruction task. We found that directly applying Mamba for 3D hand reconstruction results in inaccurate meshes due to its unidirectional scanning and the lack of specific design for 3D hand reconstruction. To tackle this challenge, we propose a Graph-guided Bidirectional Scan (GBS) to effectively capture the semantic and spatial relation between joints, as shown in Figure~\ref{fig:motivation}(c).
Besides, the transformer's attention requires calculating correlation among all tokens and introduces unnecessary background correlations, while our proposed GBS uses 88.5\% fewer tokens (see Section~\ref{sec:hamba} for more details). Secondly, though Mamba-based models~\cite{behrouz2024graph,gu2023mamba,hu2024zigma,lieber2024jamba,wang2024graphmamba,zhang2024motion} excel in modeling long-range sequences, they are not proficient at capturing the local-relation information (in our case, the `semantics' between hand joints). Since graph learning has the capability to effectively capture node relations, we integrate graph convolutions into state space modeling, significantly enhancing the representation by considering the intricate hand joint relations.

In particular, to effectively leverage state space modeling (SSM) and graph learning capabilities for 3D hand reconstruction, we first carefully design a Token Sampler (TS) under guidance with hand joints predicted by Joint Regressor (JR), then feed sampled token into the Graph-guided State Space block (GSS) under the Graph-guided Bidirectional Scan (GBS). Lastly, we introduce a fusion module to integrate the state space tokens and global features to further improve performance. As shown in Figure~\ref{fig:teaser}, Hamba achieves significant visual performance in challenging scenarios. We summarize our contributions as follows:
\vspace{-0.5em}
\setlist{leftmargin=4mm}
\begin{itemize}
    \item We are the \textit{first} to incorporate graph learning and state space modeling (SSM) for reconstructing robust 3D hand mesh. Our key idea is to reformulate the Mamba scanning into graph-guided bidirectional scanning for 3D reconstruction using a few effective tokens.
    \item We propose a simple yet effective Graph-guided State Space (GSS) block to capture structured relations between hand joints using graph convolution layers and Mamba blocks.
    \item We introduce a token sampler that effectively boosts performance. A fusion module is also introduced to further enhance performance by integrating state space tokens and global features.
    \item Extensive experiments on multiple challenging benchmarks demonstrate Hamba's superiority over current SOTAs, achieving impressive performance for in-the-wild scenarios.
\end{itemize}
\vspace{-0.75em}

\section{Related Works}
\label{Related Works}
\vspace{-0.5em}

\noindent \textbf{3D Hand Reconstruction.} Multiple approaches have been proposed to reconstruct 3D hand mesh~\cite{ballan2012motion,khamis2015learning,melax2013dynamics,oberweger2015training,oikonomidis2011efficient,schmidt2014dart,sridhar2013interactive,
tkach2016sphere,tzionas2016capturing}, with most works leveraging the MANO~\cite{romero2022MANO} parametric representation of the 3D hand. Zhang~\textit{et al.}~\cite{zhang2019HAMR} utilized a CNN encoder to iteratively regress the hand mesh based on heatmaps under 2D, 3D, silhouette, and geometric constraints. I2L-MeshNet~\cite{moon2020i2l} proposed line pixel-based 1D heatmaps for estimating joint locations and regressing MANO parameters, while HandAR~\cite{tang2021handAR} estimated parameters through three stages: joint, mesh, and a refining stage to combine previous features. The joint stage applies a multitask decoder to predict both hand joints and the segmentation mask. MeshGraphormer~\cite{lin2021mesh} introduced a graph residual block into the transformer to enhance the spatial structure. HaMeR~\cite{pavlakos2024reconstructing} showed that a simple but large transformer-based architecture trained on a large dataset can achieve SOTA performance. SimpleHand~\cite{zhou2024simple} sampled tokens with UV predictions on a high-resolution feature map, cooperating with a cascade upsampling decoder. They further compare different combinations of token generation strategies are compared, including global feature, grid sampling, keypoint sampling, $4\times$ upsampling feature map, and coarse-mesh-guided point sampling. Recently, HHMR~\cite{li2024hhmr} proposed a graph diffusion model to learn a prior of gestures and inpaint the occluded hand portion. To further enhance performance, we propose the graph-guided state space model to leverage joint relations and capture spatial joint sequences.

\noindent \textbf{State Space Models (SSMs).} State space was originally elaborated in Kalman filtering~\cite{kalman1960new} that described states and transitions with first-order differential equations. Structured State Space Sequence (S4) models~\cite{gu2021s4,gu2021combining} have the capability to model dependencies. Recently, Mamba~\cite{gu2023mamba} further improved the S4 models by expanding their fixed projection matrices linearly with the input sequence length. Many recent works have adapted Mamba for visual learning, leveraging its global receptive field and dynamic weights. Liu \textit{et al.}~\cite{liu2024vmamba} and Yang \textit{et al.}~\cite{vim} used Mamba for classification, segmentation, and object detection tasks. To effectively capture the spatial relations, they scanned the input image patches forward and backward horizontally. VMamba~\cite{liu2024vmamba} further added two vertical directions creating a cross-scan. Zhang~\textit{et al.}~\cite{zhang2024motion} designed a mamba model for motion generation, scanning unidirectionally along the temporal sequence and bidirectionally along channel dimensions in a hierarchy. Behrouz~\textit{et al.}~\cite{behrouz2024graph} and Wang~\textit{et al.}~\cite{wang2024graph} designed Graph-mamba to address traditional graph representation learning tasks, enhancing long-range context learning using Mamba blocks. Hamba makes the first attempt to adapt Mamba and graph learning to solve 3D hand reconstruction.

\vspace{-0.5em}
\section{Proposed Methodology}
\label{method}
\vspace{-0.5em}

\begin{figure}
  \centering
  \includegraphics[width=\textwidth]{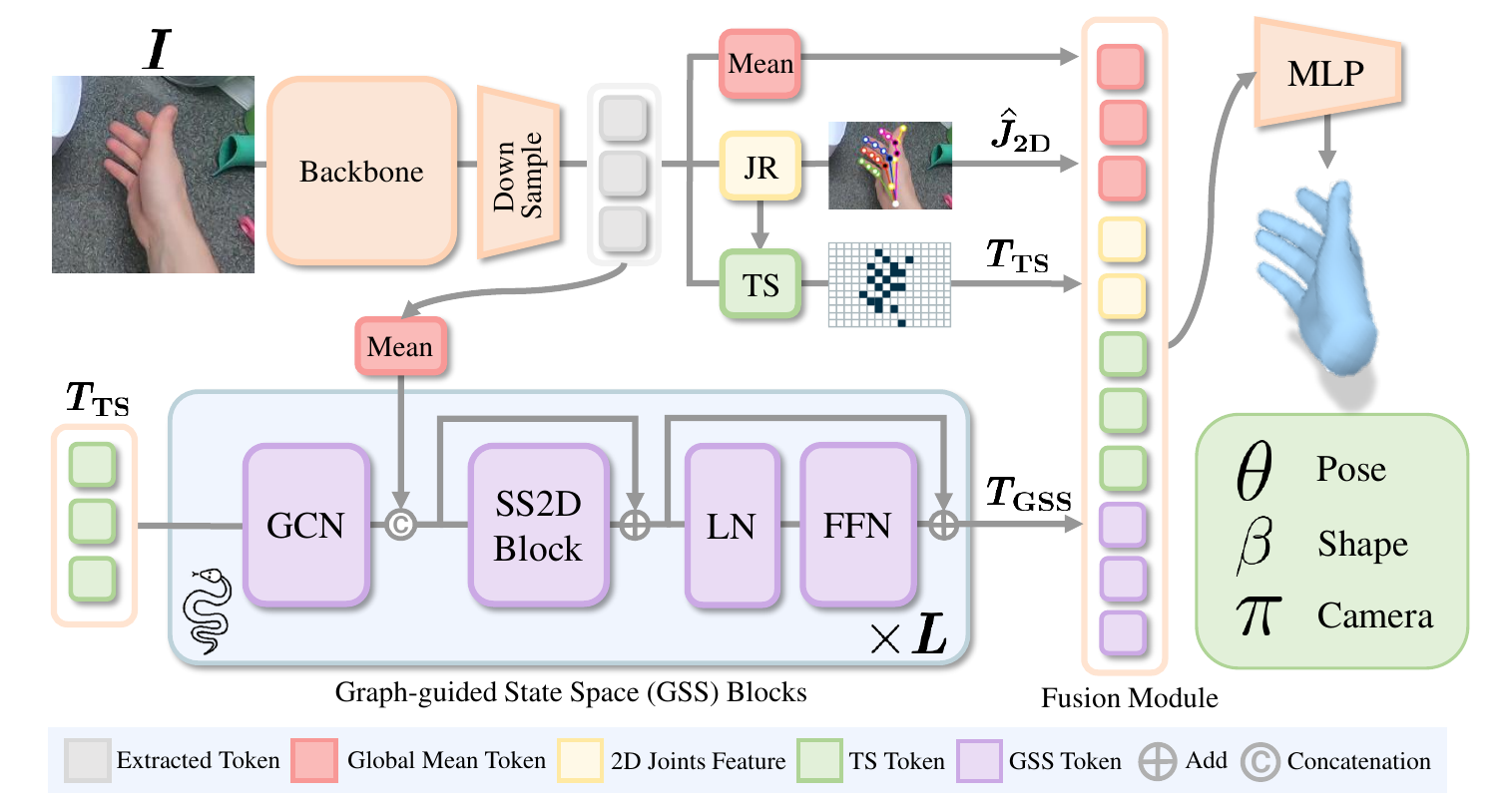}
  \vspace{-6mm}
  \caption{\textbf{Overview of Hamba's architecture}. Given a hand image $I$, tokens are extracted via a trainable backbone model and downsampled. We design a graph-guided SSM as a decoder to regress hand parameters. The hand joints ($J_{\text{2D}}$) are regressed by Joints Regressor (JR) and fed into the Token Sampler (TS) to sample tokens ($T_{\text{TS}}$). The joint spatial sequence tokens ($T_{\text{GSS}}$) are learned by the Graph-guided State Space (GSS) blocks. Inside each GSS block, the GCN network takes $T_{\text{TS}}$ as input and its output is concatenated with the mean down-sampled tokens. GSS leverages graph learning and state space modeling to capture the joint spatial relations to achieve robust 3D reconstruction.}
  \label{fig:pipeline}
  \vspace{-4mm}
\end{figure}

We propose a novel Mamba-based method that incorporates graph learning and state space modeling to learn the joint relations from the joint spatial sequence (Figure~\ref{fig:pipeline}). First, we introduce the concept of state space models (SSMs). Next, we provide the detailed principle of the proposed Token Sampler (TS), Graph-guided Bidirectional Scan (GBS), and Graph-guided State Space (GSS) modules.

\subsection{Preliminaries}

\noindent \textbf{S6 Models.} Selective Scan Structured State Space Sequence (S6) models~\cite{gu2023mamba} is a category of sequence models that have demonstrated superior ability in handling sequences. These models are primarily an extension of the previously proposed S4 models~\cite{gu2021s4}, which maps an input sequence $x(t)\in\mathbb{R} \rightarrow y(t) \in \mathbb{R}$, through the latent state $h(t) \in \mathbb{R}^N$, following ordinary linear differential equations (Eq.~\ref{eq:ssm}), where $\boldsymbol{A} \in \mathbb{R}^{N \times N}$, $\boldsymbol{B} \in \mathbb{R}^{N \times 1}$, $\boldsymbol{C} \in \mathbb{R}^{1 \times N}$ and ${D} \in \mathbb{R}^{1}$ are the weighting parameters.\\
\begin{minipage}{0.5\textwidth}
\begin{equation}
\label{eq:ssm}
    \begin{split}
  &h'(t) = \boldsymbol{A}h(t) + \boldsymbol{B}x(t),  \\
  &y(t)  = \boldsymbol{C}h(t)+ {D}x(t),
\end{split}
\end{equation}
\end{minipage}
\begin{minipage}{0.5\textwidth}
\vspace{0.5em}
\begin{equation}
\begin{aligned}
    &h_{t} = \overline{\boldsymbol{A}}h_{t-1}+\overline{\boldsymbol{B}}x_{t},\\
    &y_{t} = \boldsymbol{C}h(t).
\end{aligned}\label{discretized}
\end{equation}
\end{minipage}

For practical computation, these continuous dynamical systems are discretized (Eq.~\ref{discretized}). This is achieved by using the zero-order hold (ZOH) discretization rule (Eq.~\ref{ZOH}).
\begin{align}
    \overline{\boldsymbol{A}}=\text{exp}(\Delta\boldsymbol{A}), \;\;\; \overline{\boldsymbol{B}}=(\Delta\boldsymbol{A})^{-1}(\text{exp}(\Delta\boldsymbol{A})-\boldsymbol{I})\cdot\Delta\boldsymbol{B}, 
\label{ZOH}
\end{align}
where $\Delta$ represents the discrete step size. Since both the weighting parameters and discretizing rules are fixed over time, S4 models can be viewed as linear time invariance systems. Mamba~\cite{gu2023mamba} further expands S4 models' projection matrices to scan the entire input sequence through a selective scan.

\noindent \textbf{Mamba for Visual Representation.} Since Mamba~\cite{gu2023mamba} is primarily designed for 1D data, it is challenging to directly apply it to image data with global spatial context and local relation information. Recent works~\cite{liu2024vmamba, zhu2024vision} have extended Mamba for learning visual representations. VMamba~\cite{liu2024vmamba} developed a 2D selective scan (SS2D) block and integrated it into the VSS Block (Figure~\ref{fig:blocks}(b)). The VSS block is then stacked consecutively with convolution layers for downsampling image patches via patch merging \cite{liu2022swin}. The main difference between Mamba and VSS~\cite{liu2024vmamba} block (Figure~\ref{fig:blocks}(a-b)) is replacing the S6 block with SS2D to adapt selective scanning for image data.

\subsection{Hamba}
\label{sec:hamba}

\noindent \textbf{Problem Formulation.} Given a single hand image $I$, our goal is to reconstruct the 3D hand mesh. We learn the mapping function $f(I)= \{\theta, \beta, \pi\}$ that regresses MANO~\cite{romero2022MANO} parameters from the image $I$, where $\theta \in \mathbb{R}^{48}$, $\beta \in \mathbb{R}^{10}$, and $\pi \in \mathbb{R}^{3}$ represent the pose, shape, and camera parameters, respectively. Finally, the MANO model $\mathcal{M}(\theta, \beta)$ generates the corresponding hand mesh $M\in\mathbb{R}^{778 \times 3}$. 

\noindent \textbf{Model Architecture.} Figure~\ref{fig:pipeline} illustrates the Hamba model architecture. First, we feed the hand image $I\in \mathbb{R}^{H \times W \times 3}$ into a ViT~\cite{dosovitskiy2020ViT,xu2022vitpose} backbone to extract tokens $T \in \mathbb{R}^{\frac{H}{16} \times \frac{W}{16} \times 1280}$ where $H$=$256$ and $W$=$192$. Tokens from the backbone are downsampled from dimensions 1280 to 512 using convolution layers (Conv2D). Second, we sample effective tokens using a Token Sampler (TS), which utilizes the 2D joint locations predicted by a Joints Regressor (JR). These tokens are fed into the Graph-guided State Space (GSS) block, which exploits the joint spatial sequence by modeling its state space using the proposed Graph-guided Bidirectional scan (GBS). Finally, we fuse the GSS tokens with the global mean feature, the sampled tokens, and the 2D joint features via a fusion module. Lastly, the MANO parameters are regressed using MLP.

\noindent \textbf{Token Sampler (TS) and Joints Regressor (JR).} To prevent the GSS Block from being influenced by the background and unnecessary features during the early stages of the training, it is important to select effective tokens that encode the relations between hand joints. We propose a Token Sampler (TS), which selects effective tokens utilizing the initial 2D hand joint prediction from the Joints Regressor (JR). While it is possible to use off-the-shelf 2D joint estimator like OpenPose~\cite{cao2019openpose} or MediaPipe~\cite{lugaresi2019mediapipe}, this would increase model complexity. Previous works~\cite{ren2023DIR,zhou2024simple} primarily used Conv-Pooling-FC schemes for initial joints regression. In our work, the JR consists of stacked SS2D blocks followed by an MLP head which regresses the initial MANO parameters $\{\hat{\theta}, \hat{\beta}, \hat{\pi}\}$. After the JR regresses 3D joints $\hat{J}_{\text{3D}} \in \mathbb{R}^{21 \times 3}$, these are projected back to the 2D image plane using perspective projection $\Pi$ with the predicted camera translation $\hat{\pi}$ to obtain $\hat{J}_{\text{2D}} \in \mathbb{R}^{21 \times 2}$. We denote a predefined focal length $F_{\text{focal}}=5000$ mm. Those are formulated as,
\begin{align}
    \hat{\theta}, \hat{\beta}, \hat{\pi}={\text{JR}}(T), \;\;\;
    \hat{J}_{\text{3D}} = \text{MANO}\ (\hat{\theta}, \hat{\beta}), \;\;\;
    \hat{J}_{\text{2D}} = \Pi\ (\hat{J}_{\text{3D}},F_{\text{focal}},\hat{\pi}).
\end{align}
To align the sampled tokens with 2D joints, we use bilinear interpolation. The sampled token $T_{\text{TS}} \in \mathbb{R}^{C \times J}$ is formulated as,
\begin{align}
    &T_{\text{TS}} = \text{TS}(\text{Conv2D}(T), \hat{J}_{\text{2D}}),
\end{align}
% \vspace{-1em}
where $J$ denotes the total of 21 joints, and $C$ is the token dimension of 512. 

\begin{figure}
  \centering
  \includegraphics[width=0.85\textwidth]{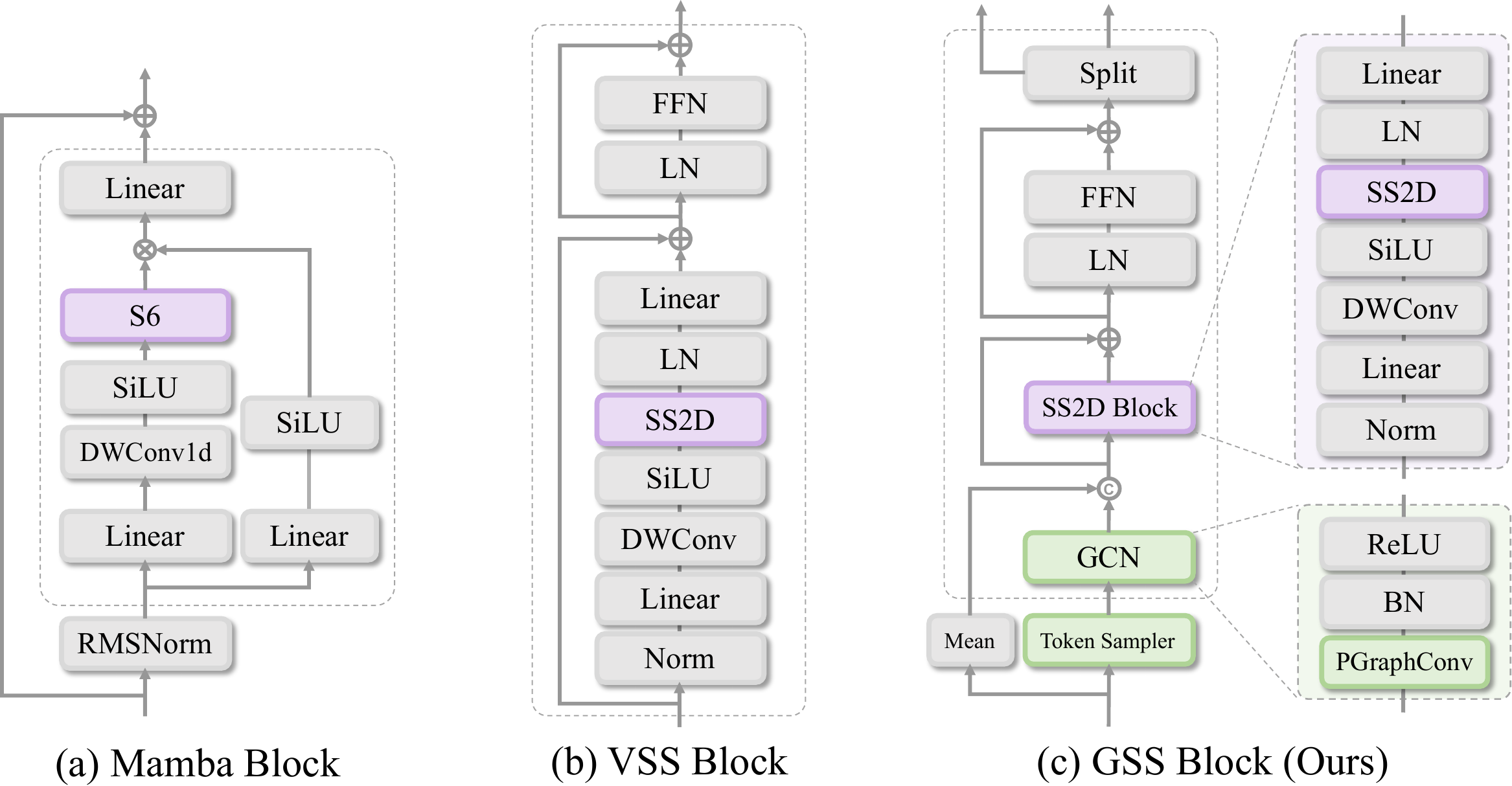}
  \vspace{-1mm}
  \caption{The illustration of the proposed Graph-guided State Space (GSS) block.}
  \label{fig:blocks}
  \vspace{-4mm}
\end{figure}

\noindent\textbf{Graph-guided Bidirectional Scan (GBS).} To achieve robust reconstruction and leverage effective tokens, we reformulate Mamba's unidirectional scanning as a graph-guided bidirectional scan, thus adapting it for 3D reconstruction tasks. GBS is designed to follow a specific graph pattern, considering the spatial and topological connection of the hand joints with image features. A naive approach would be scanning all tokenized image patches (Figure~\ref{fig:motivation}(b)). However, this involves redundant tokens, making it challenging to learn joint spatial relations effectively. To address this, we propose two novel ideas. First, instead of scanning all tokens unidirectionally, we perform hand joint-level bidirectional scanning of sampled tokens $T_{\text{TS}}$. This effectively reduces the number of tokens to be scanned from 192 to 22 ($\approx$ 88.5\% reduction). We adapt VMamba~\cite{liu2024vmamba}'s SS2D block for bidirectional scanning to be suitable for our joint spatial sequence. Second, to capture the local and global joint relations, we introduce a Semantic GCN block~\cite{zhao2019semantic}. Mamba learns long-range dependencies, but it is less effective at capturing fine-grained local information in intricate structures like the 3D mesh. The GCN learns input-independent weight matrix to model the edges between hand joints, reflecting how one joint influences another based on prior embedded in graph structures. Introducing graph learning makes it possible to explicitly encode the graph structure within our GSS module. Let $\mathcal{G} = {\mathbf{\{V, E\}}}$ be the graph, $\mathbf{V}$ is the set of $J$ nodes and $\mathbf{E}$ are the edges. $T_{\text{GCN}_l}$ represents the output of the $l$-th GCN block, while the complete output of the GSS block is $T_{\text{GSS}_l}$. For a graph-based propagation, we multiply the token with a learnable parameter matrix $\mathbf{W} \in \mathbb{R}^{C \times C}$. Thus, the GCN operation is formulated as,
\begin{equation}
    T_{\text{GCN}_{l}} = \begin{cases}
    \sigma (\mathbf{W} ~T_{\text{TS}} ~\mathbf{P}_i (\mathbf{M} \odot \mathbf{G})), &\text{\textit{l} = 1},\\
    \sigma (\mathbf{W} ~T_{\text{GSS}_{l-1}}^{\{1,..,21\}} ~\mathbf{P}_i (\mathbf{M} \odot \mathbf{G})), &\text{\textit{l} > 1},
    \end{cases}
\end{equation}
where $\mathbf{M} \in \mathbb{R}^{J \times J}$ is the learnable weighting matrix, $\mathbf{P}_i$ denotes the softmax non-linearity that is applied to normalize the input matrix for all node $i$ choices, while $\mathbf{G} \in [0, 1]^{J \times J}$ denotes the adjacency matrix of graph $\mathcal{G}$ and $\odot$ denotes element-wise multiplication. $J$ denotes the total of 21 joints, and $C$ is the token dimension of 512. 

\noindent \textbf{Graph-guided State Space (GSS) Block.} Overall, our decoder consists of $L$ GSS blocks. The GSS architecture is illustrated comparatively in Figure~\ref{fig:blocks}. In the first GSS block, the sampled tokens $T_{\text{TS}}$ are passed through graph convolution (GCN) layers. The GCN layer consists of a PGraphConv~\cite{lee2018higherorder}, a Batch Norm, and a ReLU activation. For the GCN, the adjacency matrix is defined based on the hand joint skeleton in the joint order of MANO. To provide the global context, the output from the GCN is concatenated with the global mean token along the joint token sequence. This global mean token is the mean of the downsampled image tokens. This concatenated sequence $T_{\text{GCN}_l}^{c}$ is then fed into the SS2D block and summed with the output through a residual connection. The SS2D block is followed by a Layer Norm (LN), a Feed-Forward Network (FFN), and another residual connection. For subsequent GSS blocks {\footnotesize $l \in \{2, .., L\}$}, the input is the output from the previous block {\footnotesize $T_{\text{GSS}_{l-1}}$}. Before this sequence passes through its GCN layer, it is split, and only the first 21 tokens {\footnotesize $T_{\text{GSS}_{l-1}}^{\{1,..,21\}}$} are fed to the GCN. The global mean token {\footnotesize $T_{\text{GSS}_{l-1}}^{\{22\}}$} is concatenated back with the GCN's output before it enters the SS2D block as shown in Eq.~\ref{GSS_GCN_equation} below:
\vspace{-2pt}
\begin{equation}
    T_{\text{GCN}_{l}}^{c}= \begin{cases}
    T_{\text{GCN}_{l}} \oplus \text{Mean}(\text{Conv2D}(T)), &\text{\textit{l} = 1},\\
    T_{\text{GCN}_{l}} \oplus T_{\text{GSS}_{l-1}}^{\{22\}}, &\text{\textit{l} > 1},
    \end{cases}
\label{GSS_GCN_equation}
\end{equation}
\vspace{-1em}
\begin{align}
    T_{\text{GSS}_{l}}= \text{FFN}(\text{LN}(\text{SS2D} (T_{\text{GCN}_{l}}^{c}) + T_{\text{GCN}_{l}}^{c})) + \text{SS2D}(T_{\text{GCN}_{l}}^{c}) + T_{\text{GCN}_{l}}^{c}.
\end{align}
where  $\oplus$ denotes concatenation. The GSS block not only leverages features from state space modeling and graph learning but also considers global features. This design enables Hamba to learn effective features to enhance performance by incorporating state space modeling and graph learning with few tokens, shown in our ablation study Section \ref{sec:ablation}.

\noindent\textbf{State Space Modeling for Joint Spatial Sequence.} Different from the video-based Mamba models~\cite{chen2024video,gao2024matten,li2024videomamba}, which learns the temporal feature with the frame sequence, Hamba focuses on the joint spatial sequence per frame and reveals that modeling joint relations with Mamba~\cite{gu2023mamba} can significantly improve the 3D reconstruction performance. In particular, as shown in Eq.~\ref{eq:ssm}, $x(t)$ represents $t$-th token of the joint spatial sequence, which is first sampled by the TS using the JR and then encoded with the GCN. Note that $t$ denotes the index of the hand joint iteration. Lastly, $y(t)$ is the updated token of the $t$-th of the joint spatial sequence after passing through GSS Blocks. The proposed GSS block effectively enhances 3D reconstruction performance by learning the joint spatial sequence relations with graph learning and state space modeling.

\noindent\textbf{Loss Functions.} Following~\cite{pavlakos2024reconstructing}, we train Hamba using a combined loss which includes 2D joint loss~$\mathcal{L}_{\text{2D}}$, 3D joint loss~$\mathcal{L}_{\text{3D}}$, pose loss~$\mathcal{L}_{\theta}$, shape loss~$\mathcal{L}_{\beta}$, and an adversarial loss~$\mathcal{L}_{\text{adv}}$. $\mathcal{L}_{\text{2D}}$ and $\mathcal{L}_{\text{3D}}$ are calculated using the L1 Norm, while $\mathcal{L}_{\theta}$ and $\mathcal{L}_{\beta}$ use the L2 Norm. The training loss $\mathcal{L}_{\text{total}}$ is defined as Equation~\ref{eq:loss_fn}, where $\lambda_\text{2D},~\lambda_\text{3D},~\lambda_\theta,~\lambda_\beta$, and $\lambda_\text{adv}$ denote each term's weight respectively.
% \vspace{-3mm}
\begin{align}
    \mathcal{L}_\text{total} = \lambda_\text{2D}\mathcal{L}_\text{2D} + \lambda_\text{3D}\mathcal{L}_\text{3D} + \lambda_\theta\mathcal{L}_\theta + \lambda_\beta\mathcal{L}_\beta + \lambda_\text{adv}\mathcal{L}_\text{adv},
\label{eq:loss_fn}
\end{align}
\vspace{-2.5em}

\section{Experiments}
\vspace{-2mm}

\noindent \textbf{Datasets.} We train Hamba on 2.7M training samples from multiple datasets (same setting as~\cite{pavlakos2024reconstructing} for a fair comparison) that had either both 2D and 3D hand annotations or just 2D annotations. This included FreiHAND~\cite{zimmermann2019freihand}, HO3D~\cite{hampali2020honnotate}, MTC~\cite{xiang2019monocular}, RHD~\cite{zimmermann2017learning}, InterHand2.6M~\cite{moon2020interhand2}, H2O3D~\cite{hampali2020honnotate}, DexYCB~\cite{chao2021dexycb}, COCO-Wholebody~\cite{jin2020whole}, Halpe~\cite{fang2022alphapose}, and MPII NZSL~\cite{simon2017hand} datasets.

\noindent \textbf{Implementation Details.} We set learning rate as $10^{-5}$, weight decay factor as $10^{-4}$, with the `sum' loss. Weights for each term in the loss function are $\lambda_{3D}=0.05$ for 3D keypoint loss, %($\mathcal{L}_{3D}$), 
$\lambda_{2D}=0.01$ for 2D keypoint loss, %($\mathcal{L}_{2D}$), 
$\lambda_\theta=0.001$ for global orientation and hand pose loss. Weights for beta and adversarial loss, %($\mathcal{L}_\beta, \mathcal{L}_{adv}$), 
i.e., $\lambda_\beta$ and $\lambda_{adv}$ were set as 0.0005. Ablations were run for 60k steps due to computational limitations on 2.7M dataset. Additional details are included in the Appendix~\ref{Supp:Appendix}.

\noindent \textbf{Evaluation Metrics.} Following the same protocols employed in previous works~\cite{lin2021mesh,pavlakos2024reconstructing,zhou2024simple}, we used PA-MPJPE and $\text{AUC}_J$ as the metrics for evaluating the reconstructed 3D joints and PA-MPVPE, $\text{AUC}_V$, F@5mm, and F@15mm for evaluating the reconstructed 3D mesh vertices.

\subsection{Main Results}

\noindent \textbf{3D Joints and Mesh Reconstruction Evaluation.} We test Hamba on 3 widely used benchmarks: FreiHAND~\cite{zimmermann2019freihand}, HO3Dv2~\cite{hampali2020honnotate}, and HO3Dv3~\cite{hampali2021ho}. The quantitative comparison with state-of-the-art 3D hand reconstruction models is presented in Table~\ref{tab:freihand_results}, Table~\ref{tab:HO3Dv2results}, and Table~\ref{tab:HO3Dv3} respectively. Since almost all previous methods (including the popular MobRecon~\cite{chen2022mobrecon}, MeshGraphormer~\cite{lin2021mesh}, and the recent HHMR~\cite{li2024hhmr}, SimpleHand~\cite{zhou2024simple}) were trained only using the FreiHAND~\cite{zimmermann2019freihand}, for a fair comparison, we compared them with the Hamba version trained using only the FreiHAND~\cite{zimmermann2019freihand} dataset. Meanwhile, for a fair comparison with HaMeR~\cite{pavlakos2024reconstructing}, we trained Hamba on the same datasets as HaMeR~\cite{pavlakos2024reconstructing} for all other comparisons. Many methods, including the popular MeshGraphormer~\cite{lin2021mesh} and METRO~\cite{lin2021end}, report their metrics using Test-Time Augmentation (TTA) which boosts the final results. We report our performances, both with and without TTA. In both scenarios, Hamba significantly achieves better results, outperforming SOTAs in all benchmarks.

\noindent \textbf{In-the-wild Generalizability Evaluation.} Approximately $95\%$ of datasets used for training previous models~\cite{chen2022mobrecon,cho2022FastMETRO,li2024hhmr,lin2021end,lin2021mesh,pavlakos2024reconstructing,vasu2023fastvit,zhou2024simple} were collected in controlled indoor environments, such as studios or multi-camera setups. This includes the FreiHAND~\cite{zimmermann2019freihand}, HO3Dv2~\cite{hampali2020honnotate}, and HO3Dv3~\cite{hampali2021ho} benchmarks that are popularly used for both training and evaluation. However, training models on datasets collected in controlled environments often leads to decreased performance in real-world scenarios. Thus, solely evaluating performance over indoor-collected datasets might not provide a correct evaluation of the robustness of 3D hand reconstruction. We additionally evaluate Hamba's in-the-wild performance on the recently proposed HInt~\cite{pavlakos2024reconstructing} benchmark, which has variations in visual conditions, viewpoints, and hand interactions. Since HInt-NewDays~\cite{cheng2023towards} and HInt-EpicKitchensVISOR~\cite{damen2018scaling,VISOR2022} annotations are 2D keypoints, PCK~\cite{yang2012articulated} computed at varying thresholds is used as the evaluation metrics. As shown in Table~\ref{tab:hint}, Hamba outperforms existing models by a large margin and surpasses HaMeR~\cite{pavlakos2024reconstructing}, showing improvement in model robustness for in-the-wild scenarios. None of the models (including Hamba) have been trained on/ previously ever seen HInt dataset.

\begin{figure}[t]
  \centering
  \includegraphics[width=\textwidth]{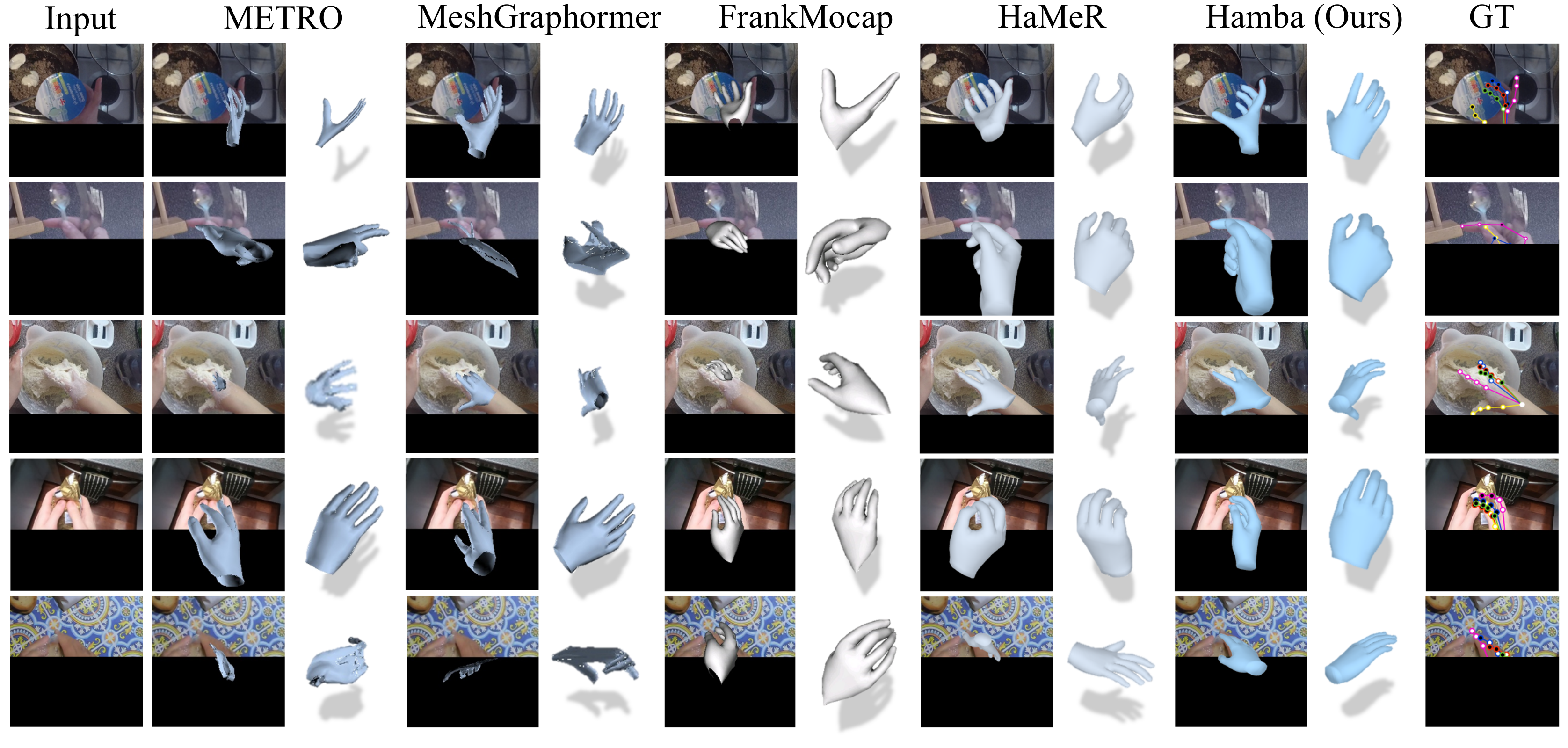}
  \vspace{-6mm}
  \caption{\textbf{Qualitative in-the-wild comparison} of the proposed Hamba with SOTAs on HInt-EpicKitchensVISOR~\cite{damen2018scaling,pavlakos2024reconstructing}. None of the models (including Hamba) have been trained on HInt.}
  \label{fig:compare}
  \vspace{-3mm}
\end{figure}

\noindent \textbf{Qualitative Comparison.} Figure~\ref{fig:compare} presents the qualitative comparison of Hamba's 3D hand mesh reconstruction with SOTA models on in-the-wild images from HInt-EpicKitchens. This includes models that directly regress vertices (METRO~\cite{lin2021end}, MeshGraphormer~\cite{lin2021mesh}), and parametric methods, which regress MANO parameters (FrankMocap~\cite{rong2021frankmocap}, HaMeR~\cite{pavlakos2024reconstructing}). These images are particularly challenging since they comprise real-world cooking videos of a person with highly occluded hands, hand-hand, and/or hand-object interactions. For visual comparison, we select images where the hand lies in the corners, causing a truncation scenario thus increasing the complexity further. Hamba consistently outperforms other models and achieves a much better reconstruction. From Figure~\ref{fig:compare}, we can observe that in severe in-the-wild truncation scenarios, Hamba achieves better hand reconstruction, even though the hand is truncated or occluded. We attribute this performance to effectively learning the spatial hand joint sequence with the state space model. The same is verified in the ablation study presented in Sec.~\ref{sec:ablation}. Figure~\ref{fig:additional_results} presents in-the-wild results on various movies, interviews, etc., scenarios. Figure~\ref{fig:new_days_in_the_wild} and Figure~\ref{fig:VISOR_in_the_wild} presents additional visual results on HInt-NewDays and HInt-EpicKitchensVISOR respectively.  Hamba can robustly reconstruct 3D hands in various complicated hand gestures like grasping, holding, grabbing, finger-pointing, and flattening from different viewing directions, even in heavily occluded and truncated scenarios.

\begin{table}[t]
\renewcommand{\arraystretch}{0.85}
\setlength{\tabcolsep}{4pt}
    \centering
    \caption{Comparison with SOTAs on~\textbf{FreiHAND} dataset~\cite{zimmermann2019freihand}. $^*$Stacked Structure; {\color{red} $^{\dagger}$}used Test-Time Augmentation (TTA). Best scores highlighted {\footnotesize \colorbox{green!50}{Green}}, while second best are highlighted {\footnotesize \colorbox{green!20}{Light Green}}. PA-MPJPE and PA-MPVPE are measured in mm. -: Info not reported by model.}
    \label{tab:freihand_results}
    \vspace{-1.5mm}
    \resizebox{1\textwidth}{!}{
    \begin{tabular}{lcc|cccc}
    \toprule
    Method & Venue & Backbone  &  PA-MPJPE $\downarrow$ & PA-MPVPE $\downarrow$ & F@5mm $\uparrow$ & F@15mm $\uparrow$ \\
    
    \midrule
    Zimmermann~\textit{et~al}.~\cite{zimmermann2019freihand}   & ICCV~19 & ResNet50 & - & 10.7 & 0.529 & 0.935 \\
    Boukhayma~\textit{et~al.}~\cite{boukhayma20193d} & CVPR~19 & ResNet50 & - & 13.0 & 0.435 & 0.898   \\
    ObMan~\cite{hasson2019learning}   & CVPR~19 & ResNet18     &  -  & 13.2 & 0.436 & 0.908            \\
    MobileHand~\cite{lim2020mobilehand}& ICONIP~20 & MobileNet  &  -  & 13.1 & 0.439 & 0.902           \\
    YoutubeHand~\cite{kulon2020weakly}& CVPR~20 & ResNet50     & 8.4 &  8.6 & 0.614 & 0.966            \\
    Pose2Mesh~\cite{choi2020pose2mesh}& ECCV~20 & $-$          & 7.7 &  7.8 & 0.674 & 0.969            \\
    I2L-MeshNet~\cite{moon2020i2l}    & ECCV~20 & ResNet50$^*$ & 7.4 &  7.6 & 0.681 & 0.973            \\
    HIU-DMTL~\cite{zhang2021hand}     & ICCV~21 & Custom$^*$   & 7.1 &  7.3 & 0.699 & 0.974            \\
    CMR~\cite{chen2021camera}         & CVPR~21 & ResNet50$^*$ & 6.9 &  7.0 & 0.715 & 0.977            \\   
    I2UV-HandNet~\cite{chen2021i2uv}  & ICCV~21 & ResNet50     & 6.7 &  6.9 & 0.707 & 0.977            \\ 
    Tang~\textit{et al.}~\cite{tang2021towards} & ICCV~21 & ResNet50 & 6.7 &  6.7 & 0.724 & 0.981            \\   
    METRO{\color{red} $^{\dagger}$}~\cite{lin2021end} & CVPR~21 & HRNet & 6.3 &  6.5 & 0.731 & 0.984            \\
    MeshGraphormer{\color{red} $^{\dagger}$}\cite{lin2021mesh} & ICCV~21 & HRNet & 5.9 &  6.0 & 0.764 & 0.986 \\
    MobRecon~\cite{chen2022mobrecon}  & CVPR~22 & ResNet50$^*$ & \cellcolor{green!50} \textbf{5.7} &  5.8 & 0.784 & 0.986            \\    
    FastMETRO~\cite{cho2022FastMETRO} & ECCV~22 & HRNet & 6.5 &  7.1 & 0.687 & 0.983            \\    
    FastViT~\cite{vasu2023fastvit} & ICCV~23 & FastViT & 6.6 &  6.7 & 0.722 & 0.981            \\    
    AMVUR~\cite{jiang2023probabilistic}& CVPR~23& ResNet50     & 6.2 &  6.1 & 0.767 & 0.987            \\    
    Deformer~\cite{yoshiyasu2023deformable} & CVPR~23 & HRNet & 6.2 &  6.4 & 0.743 & 0.984            \\    
    PointHMR~\cite{kim2023sampling}   & CVPR~23 & HRNet        & 6.1 &  6.6 & 0.720 & 0.984            \\    
    Zhou~\textit{et~al.}~\cite{zhou2024simple} & CVPR~24 & FastViT & \cellcolor{green!50} \textbf{5.7} &  6.0 & 0.772 & 0.986   \\
    HaMeR~\cite{pavlakos2024reconstructing} & CVPR~24 & ViTPose& 6.0 &  5.7 & 0.785 & 0.990            \\
    HaMeR-170k~\cite{pavlakos2024reconstructing} & CVPR~24 & ViTPose& 6.1 &  5.8 & 0.782 & 0.990            \\
    HHMR{\color{red} $^{\dagger}$}~\cite{li2024hhmr}            & CVPR~24 & ResNet50     & \cellcolor{green!20} 5.8 & 5.8 & - & - \\
    \midrule
    \bf Hamba                         & \bf Ours    & ViTPose   & \cellcolor{green!20} 5.8  & \cellcolor{green!20} 5.5  & \cellcolor{green!20} 0.798   &  \cellcolor{green!20} 0.991 \\
    \bf Hamba{\color{red} $^{\dagger}$} & \bf Ours    & ViTPose   & \cellcolor{green!50} \textbf{5.7}  & \cellcolor{green!50} \textbf{5.3}  & \cellcolor{green!50} \textbf{0.806}   &  \cellcolor{green!50} \textbf{0.992} \\
    \bottomrule
    \end{tabular}}
    \vspace{-4mm}
\end{table}

\begin{table}[t]
\renewcommand{\arraystretch}{0.85}
     \centering
    \caption{Comparison with SOTAs on \textbf{HO3Dv2}~\cite{hampali2020honnotate} hand-object interaction benchmark. %Backbones are the same as in Table~\ref{tab:freihand_results}.}
    }
    \vspace{-1mm}
    \label{tab:HO3Dv2results}
    \setlength{\tabcolsep}{4pt}
    \resizebox{1\textwidth}{!}{
    \begin{tabular}{lc|cccccc}
    \toprule
    % \hline
    Method & Venue & PA-MPJPE$\downarrow$ & PA-MPVPE$\downarrow$ & F@5mm$\uparrow$ & \ \ \ \ F@15mm$\uparrow$ & \ \ \ \ AUC$_J\uparrow$ & \ \ \ \ AUC$_V\uparrow$ \\
        
    \midrule
    ObMan~\cite{hasson2019learning}                  & CVPR~19          & 11.0  & 11.2 & 0.464 & 0.939 & 0.780 & 0.777 \\
    Pose2Mesh~\cite{choi2020pose2mesh}               & ECCV~20           & 12.5  & 12.7 & 0.441 & 0.909 & 0.754 & 0.749 \\
    I2L-MeshNet~\cite{moon2020i2l}                   & ECCV~20 & 11.2  & 13.9 & 0.409 & 0.932 & 0.775 & 0.722 \\
    Hampali~\textit{et~al}.~\cite{hampali2020honnotate} & CVPR~20       & 10.7  & 10.6 & 0.506 & 0.942 & 0.788 & 0.790 \\
    S2Hand~\cite{chen2021model}                      & CVPR~21        & 11.4  & 11.2 & 0.450 & 0.930 & 0.773 & 0.777\\
    METRO~\cite{lin2021end}                          & CVPR~21   & 10.4 & 11.1 & 0.484 & 0.946 & 0.792 & 0.779 \\
    Liu~\textit{et~al}.~\cite{liu2021semi}           & CVPR~21         & 9.9  & 9.5  & 0.528 & 0.956 & 0.803 & 0.810 \\
    I2UV-HandNet~\cite{chen2021i2uv}                 & ICCV~21          & 9.9  & 10.1 & 0.500 & 0.943 & 0.804 & 0.799 \\
    Tse~\textit{et~al.}~\cite{tse2022collaborative}  & CVPR~22          & -    & 10.9 & 0.485 & 0.943 & -    & -     \\
    ArtiBoost~\cite{yang2022artiboost}               & CVPR~22          & 11.4 & 10.9 & 0.488 & 0.944 & 0.773 & 0.782 \\
    KPT-Transf~\cite{hampali2022keypoint}       & CVPR~22          & 10.8 & -    & - & - & 0.786 & -             \\
    MobRecon~\cite{chen2022mobrecon}                 & CVPR~22          & 9.2  & 9.4  & 0.538 & 0.957 & -    & -      \\
    HandOccNet~\cite{park2022handoccnet}             & CVPR~22          & 9.1  & 8.8  & 0.564 & 0.963 & 0.819  & 0.819 \\
    HFL-Net~\cite{lin2023harmonious}                 & CVPR~23          & 8.9 & 8.7 & 0.575 & 0.965 & - & -\\
    % HandGCAT~\cite{wang2023handgcat} & ICME~23 & 8.7 &  8.7 &  0.584 & 0.963 & 0.826 & 0.827 \\
    H2ONet~\cite{xu2023h2onet}                       & CVPR~23          & 8.5 & 8.6 & 0.570 & 0.966 & 0.829 & 0.828    \\
    AMVUR~\cite{jiang2023probabilistic}              & CVPR~23          & 8.3  & 8.2  & 0.608 & 0.965 & 0.835 & 0.836 \\
    HOISDF~\cite{qi2024hoisdf} & CVPR~24 & 9.2 & - & - & - & - & - \\ 
    % SPMHand~\cite{lu2024spmhand} & TMM~24 & 8.5 &  8.5 &  0.579 & 0.968 & - & - \\
    HandBooster~\cite{xu2024handbooster} & CVPR~24 & 8.2 & 8.4 & 0.585 & 0.972 & 0.836 & 0.832 \\
    HaMeR~\cite{pavlakos2024reconstructing}          & CVPR~24 &  7.7  & \cellcolor{green!20} 7.9  &  0.635 &  0.980 &  0.846 &  0.841 \\
    HaMeR-170k~\cite{pavlakos2024reconstructing}          & CVPR~24 & \cellcolor{green!20} 7.6  & \cellcolor{green!20} 7.9  & \cellcolor{green!20} 0.639 & \cellcolor{green!20} 0.981 & \cellcolor{green!20} 0.848 & \cellcolor{green!20} 0.843 \\
    \midrule
    \textbf{Hamba}                                   & \textbf{Ours}  &  \cellcolor{green!50} \textbf{7.5}    & \cellcolor{green!50} \textbf{7.7}  &  \cellcolor{green!50} \textbf{0.648}    & \cellcolor{green!50}\textbf{0.982} &  \cellcolor{green!50}\textbf{0.850} & \cellcolor{green!50}\textbf{0.846}      \\ 
    \bottomrule
    \end{tabular}
    }
\end{table}

\begin{table}
\renewcommand{\arraystretch}{0.80}
\caption{Evaluation on \textbf{HO3Dv3}~\cite{hampali2021ho} benchmark. We only list SOTAs that reported on HO3Dv3.}
  \label{tab:HO3Dv3}
  \vspace{-1mm}
  \centering
  \resizebox{1\linewidth}{!}{
    \begin{tabular}{lc|cccccc}
    \toprule    
    Method \ \ \ \ \ \ \ \ \ \ \ \ \ \ \ \ \ \ & Venue & PA-MPJPE$\downarrow$ & PA-MPVPE$\downarrow$ & F@5mm$\uparrow$ & \ \ \ \ F@15mm$\uparrow$ & \ \ \ \ AUC$_J\uparrow$ & \ \ \ \ AUC$_V\uparrow$\\
    \midrule
    S$^2$HAND~\cite{chen2021model} & CVPR~21 & 11.5 & 11.1 & 0.448 & 0.932 & 0.769 & 0.778\\
    KPT-Transf.~\cite{hampali2022keypoint} & CVPR~22 & 10.9 & - & - & - & 0.785 & - \\
    ArtiBoost~\cite{yang2022artiboost} & CVPR~22 & 10.8  & 10.4 & 0.507 & 0.946 & 0.785 & 0.792 \\
    Yu~\textit{et al.}~\cite{yu2022uv} & BMVC~22 & 10.8 &  10.4 &  - & - & - & - \\
    HandGCAT~\cite{wang2023handgcat} & ICME~23 & 9.3  & 9.1 & 0.552 & 0.956 & 0.814 & 0.818 \\
    AMVUR~\cite{jiang2023probabilistic} & CVPR~23 & \cellcolor{green!20} 8.7 &  \cellcolor{green!20} 8.3 &  \cellcolor{green!20} 0.593 &  \cellcolor{green!20} 0.964 &  \cellcolor{green!20} 0.826 &  \cellcolor{green!20} 0.834 \\
    HMP~\cite{duran2024hmp} & WACV~24 &  10.1 &  - &  - & - & - & - \\
    SPMHand~\cite{lu2024spmhand} & TMM~24 & 8.8 &  8.6 &  0.574 & 0.962 & - & - \\
    \midrule
    \textbf{Hamba} & \textbf{Ours} & \cellcolor{green!50} \textbf{6.9} & \cellcolor{green!50} \textbf{6.8} & \cellcolor{green!50} \textbf{0.681} & \cellcolor{green!50} \textbf{0.982} & \cellcolor{green!50} \textbf{0.861} & \cellcolor{green!50} \textbf{0.864}\\
    \bottomrule
    \end{tabular}
}
    \vspace{-2mm}
\end{table}

\subsection{Ablation Studies}
\label{sec:ablation}

\noindent\textbf{Effect of Branch-wise Features.} We verify the effectiveness of each branch feature by excluding their respective tokens from the fusion module as shown in Table~\ref{tab:ablation}. First, we verify the contribution of the proposed GSS branch. When the GSS tokens are excluded (Row 3), we observe a major drop in model performance. Specifically, F@5mm ($\uparrow$) drops from 0.738 $\rightarrow$ 0.717, and the PA-MPJPE ($\downarrow$) and PA-MPVPE ($\downarrow$) errors increase from 6.6 $\rightarrow$ 6.9 and 6.3 $\rightarrow$ 6.6. Thus, in addition to local and global contexts, incorporating structured state-space representations can be effective for 3D hand reconstruction. Moreover, it is important to note that modeling spatial joint sequence relations provides better tokens than directly using the 2D joint locations, even though the latter has a clear semantic meaning for all the hand joints. We attribute this to cases of occlusions where the 2D joints cannot be precisely predicted.

\begin{wraptable}{r}{0.62\textwidth}
\vspace{-6mm}
% \begin{table}[t]
\renewcommand{\arraystretch}{0.80}
  \caption{\textbf{Ablation study} on \textbf{FreiHAND}~\cite{zimmermann2019freihand} to verify proposed components. All variants are trained for same number of steps. PA-MPJPE, PA-MPVPE and without are abbreviated as PJ, PV, `w/o'.}
   \vspace{1mm}
  \label{tab:ablation}
  \centering
  \resizebox{0.62\textwidth}{!}{
    \begin{tabular}{c|l|cccc}
    \toprule
    & Ablation & PJ $\downarrow$ & PV $\downarrow$ & F@5 $\uparrow$ & F@15 $\uparrow$ \\
    \midrule
    \multicolumn{2}{l}{Branch-wise} & & & & \\
    \midrule
    1 & w/o Token\_Sampler\_Branch         & 6.8 & 6.5 & 0.722 & 0.987 \\
    2 & w/o 2D\_Joints\_Feature\_Branch     & 6.8 & 6.6 & 0.718 & 0.986 \\
    3 & w/o GSS\_Token\_Branch         & 6.9 & 6.6 & 0.717 & 0.986 \\
    4 & w/o Global\_Mean\_Token\_Branch & 7.3 & 7.2 & 0.680 & 0.982 \\
    \midrule
    \multicolumn{2}{l}{Component-wise} & & & &  \\
    \midrule
    5 & w/o Token\_Sampler          & 6.8 & 6.6 & 0.717 & 0.986 \\
    6 & w/o Bidirectional\_Scan          & 6.9 & 6.6 & 0.718 & 0.986 \\
    7 & w/o GCN                         & 7.3 & 7.2 & 0.673 & 0.983 \\
    8 & w/o Graph-guided\_Bi\_Scan & 7.3 & 7.1 & 0.680 & 0.983 \\
    9 & w/o Mamba (SS2D+LN+FFN) & 7.3 & 7.2 & 0.675 & 0.983 \\
    \midrule
    \multicolumn{2}{l}{\bf Hamba (Full)} & \textbf{6.6} & \textbf{6.3}  &  \textbf{0.738}  & \textbf{ 0.988}  \\
    \bottomrule
  \end{tabular}
  }
  \vspace{-4mm}
\end{wraptable}

Removing the Token sampler (Row 1) or the 2D joints (Row 2) features also shows a performance drop, but is less significant than removing the GSS branch, since they only provide the local context while GSS tokens provide both local and spatial-relations information. Note that the Global Mean token (Row 4) remains important since it captures the global context, which is discarded in the ablation.

\noindent\textbf{Effect of proposed Components.} Since the GSS Block stands as a major contribution, we additionally evaluate the effectiveness of each component in the proposed GSS block. The same is presented in Table~\ref{tab:ablation}. The GSS block models the hand-joint topological structure, learning the graph-structured relations and spatial sequences of joints via graph and state space modeling. Adopting graph learning additionally provides the local context. Excluding the GCN, i.e., when simply using a Mamba block, the structure information will be neglected from the input to the SS2D block, which leads to a large drop in performance (Row 7). This indicates that the GCN is an essential component of the GSS block and using SS2D blocks alone does not lead to accurate 3D hand mesh reconstruction. A potential counter-argument may be that the input features and the GCN alone are sufficient for 3D hand reconstruction, without much improvement from the Mamba Blocks. We removed the Mamba blocks and the GSS degenerates into simple GCN, leading to an equal performance drop (Row 9). Specifically, the PA-MPJPE ($\downarrow$) and PA-MPVPE ($\downarrow$) increase from 6.6 $\rightarrow$ 7.3 and 6.3 $\rightarrow$ 7.2 respectively, while the F@5mm ($\uparrow$) and F@15mm ($\uparrow$) drop from 0.738 $\rightarrow$ 0.675 and 0.988 $\rightarrow$ 0.983 respectively. This confirms that both the GCN and the Mamba blocks are equally important in the GSS Block. To verify the effectiveness of the bidirectional scanning, we replaced it with conventional unidirectional scanning to compare, denoted as w/o Bidirectional-scan (Row 6), and the reconstruction error increased. An even larger drop in performance is observed when the proposed GBS scan is removed from the model (Row 8). When not using the token sampler, we also see a drop in performance (Row 5). Overall, Table~\ref{tab:ablation} verifies the effectiveness of each proposed component. We additionally validate this by a qualitative evaluation (in Figure~\ref{fig:ablation_visualization}), wherein the visual result gets worse when we remove the proposed modules.

\begin{table}[t]
\renewcommand{\arraystretch}{0.80}
 \caption{In-the-wild generalizability evaluation on  \textbf{HInt}~\cite{pavlakos2024reconstructing}. PCK is used as the evaluation metric.}
    \label{tab:hint}
    \centering
    \resizebox{1\textwidth}{!}{
    \begin{tabular}{@{}c|lc|ccc|ccc|ccc@{}}
        \toprule
        \multirow{2}{*}{} & \multicolumn{1}{c}{\multirow{2}{*}{Method}} & \multicolumn{1}{c}{\multirow{2}{*}{Venue}} & \multicolumn{3}{c}{NewDays} & \multicolumn{3}{c}{VISOR}  & \multicolumn{3}{c}{Ego4D}  \\
        & \multicolumn{1}{c}{} & & @0.05$\uparrow$ & @0.1$\uparrow$ & @0.15$\uparrow$  & @0.05$\uparrow$  & @0.1$\uparrow$  & @0.15$\uparrow$ & @0.05$\uparrow$  & @0.1$\uparrow$  & @0.15$\uparrow$ \\ 
        \midrule
        \parbox[t]{2mm}{\multirow{7}{*}{\rotatebox[origin=c]{90}{All Joints}}}
        & METRO~\cite{lin2021end} & CVPR~21 & 14.7 & 38.8 & 57.3 & 16.8 & 45.4 & 65.7 & 13.2 & 35.7 & 54.3 \\
        & FrankMocap~\cite{rong2021frankmocap} & ICCVW~21 & 16.1 & 41.4 & 60.2 & 16.8 & 45.6 & 66.2 & 13.1 & 36.9 & 55.8 \\
        & MeshGraphormer~\cite{lin2021mesh} & ICCV~21 & 16.8 & 42.0 & 59.7 & 19.1 & 48.5 & 67.4 & 14.6 & 38.2 & 56.0\\
        & HandOccNet (param)~\cite{park2022handoccnet} & CVPR~22 & 9.1 & 28.4 & 47.8 & 8.1 & 27.7 & 49.3 & 7.7 & 26.5 & 47.7 \\
        & HandOccNet (no param) & CVPR~22 & 13.7 & 39.1 & 59.3 & 12.4 & 38.7 & 61.8 & 10.9 & 35.1 & 58.9 \\
        & HaMeR~\cite{pavlakos2024reconstructing} & CVPR~24 & \cellcolor{green!20}48.0 & 78.0 & 88.8 & 43.0 & 76.9 & 89.3 & \cellcolor{green!20}38.9 & 71.3 & 84.4 \\ 
        & HaMeR-170k~\cite{pavlakos2024reconstructing} & CVPR~24 & 46.9 & \cellcolor{green!20}78.6 & \cellcolor{green!20}89.7 & \cellcolor{green!20}44.4 & \cellcolor{green!20}79.3 & \cellcolor{green!20}91.1 & 37.3 & \cellcolor{green!20}71.6 & \cellcolor{green!20}85.1\\ 
        & \bf Hamba & \bf Ours & \cellcolor{green!50} \textbf{48.7}  & \cellcolor{green!50} \textbf{79.2}  & \cellcolor{green!50}\textbf{90.0} & \cellcolor{green!50}\textbf{47.2} & \cellcolor{green!50}\textbf{80.2} & \cellcolor{green!50}\textbf{91.2} & \cellcolor{green!50}\textbf{41.7} & \cellcolor{green!50}\textbf{72.9} & \cellcolor{green!50}\textbf{85.5}  \\
        
        \midrule
        \parbox[t]{2mm}{\multirow{7}{*}{\rotatebox[origin=c]{90}{Visible Joints}}} 
        & METRO~\cite{lin2021end} & CVPR~21 & 19.2 & 47.6 & 66.0 & 19.7 & 51.9 & 72.0 & 15.8 & 41.7 & 60.3 \\
        & FrankMocap~\cite{rong2021frankmocap} & ICCVW~21 & 20.1 & 49.2 & 67.6 & 20.4 & 52.3 & 71.6 & 16.3 & 43.2 & 62.0 \\
        & Mesh Graphormer~\cite{lin2021mesh} & ICCV~21 & 22.3 & 51.6 & 68.8 & 23.6 & 56.4 & 74.7 & 18.4 & 45.6 & 63.2 \\
        & HandOccNet (param)~\cite{park2022handoccnet} & CVPR~22 & 10.2 & 31.4 & 51.2 & 8.5 & 27.9 & 49.8 & 7.3 & 26.1 & 48.0 \\
        & HandOccNet (no param) & CVPR~22 & 15.7 & 43.4 & 64.0 & 13.1 & 39.9 & 63.2 & 11.2 & 36.2 & 60.3 \\
        & HaMeR~\cite{pavlakos2024reconstructing} & CVPR~24 & \cellcolor{green!20}60.8 & \cellcolor{green!20}87.9 & 94.4 & 56.6 & 88.0 & 94.7 & \cellcolor{green!20}52.0 & \cellcolor{green!20}83.2 & 91.3 \\ 
        & HaMeR-170k~\cite{pavlakos2024reconstructing} & CVPR~24 & 58.1 & 87.8 & \cellcolor{green!20}94.7 & \cellcolor{green!20}57.2 & \cellcolor{green!20}88.7 & \cellcolor{green!20}95.4 & 49.6 & 82.5 & \cellcolor{green!20}91.4\\
        & \bf Hamba & \textbf{Ours} & \cellcolor{green!50}\textbf{61.2} & \cellcolor{green!50}\textbf{88.4}  & \cellcolor{green!50}\textbf{94.9} & \cellcolor{green!50}\textbf{61.4} & \cellcolor{green!50}\textbf{89.6} & \cellcolor{green!50}\textbf{95.6} & \cellcolor{green!50}\textbf{56.0} & \cellcolor{green!50}\textbf{84.3} & \cellcolor{green!50}\textbf{91.9} \\ 
        
        \midrule
        \parbox[t]{2mm}{\multirow{7}{*}{\rotatebox[origin=c]{90}{Occluded Joints}}}
        & METRO~\cite{lin2021end} & CVPR~21 & 7.0 & 23.6 & 42.4 & 10.2 & 32.4 & 53.9 & 8.1 & 26.2 & 44.7 \\
        & FrankMocap~\cite{rong2021frankmocap} & ICCVW~21 & 9.2 & 28.0 & 46.9 & 11.0 & 33.0 & 55.0 & 8.4 & 26.9 & 45.1 \\
        & MeshGraphormer~\cite{lin2021mesh} & ICCV~21 & 7.9 & 25.7 & 44.3 & 10.9 & 33.3 & 54.1 & 8.3 & 26.9 & 44.6 \\
        & HandOccNet (param)~\cite{park2022handoccnet} & CVPR~22 & 7.2 & 23.5 & 42.4 & 7.4 & 26.1 & 46.7 & 8.0 & 26.1 & 45.7 \\
        & HandOccNet (no param) & CVPR~22 &9.8 & 31.2 & 50.8 & 9.9 & 33.7 & 55.4 & 9.6 & 31.1 & 52.7 \\
        & HaMeR~\cite{pavlakos2024reconstructing} & CVPR~24 & 27.2 & 60.8 & 78.9 & 25.9 & 60.8 & 80.7 & 23.0 & 56.9 & 76.3 \\ 
        & HaMeR-170k~\cite{pavlakos2024reconstructing} & CVPR~24 & \cellcolor{green!50}\textbf{28.9} & \cellcolor{green!20}62.4 & \cellcolor{green!20}80.5 & \cellcolor{green!20}29.4 & \cellcolor{green!20}65.7 & \cellcolor{green!20}83.9 & \cellcolor{green!20}24.6 & \cellcolor{green!20}58.7 & \cellcolor{green!50}\textbf{77.7}\\
        & \bf Hamba & \textbf{Ours} & \cellcolor{green!20}{28.2}  & \cellcolor{green!50}\textbf{62.8}  & \cellcolor{green!50}\textbf{81.1}  & \cellcolor{green!50}\textbf{29.9} & \cellcolor{green!50}\textbf{66.6}  & \cellcolor{green!50}\textbf{84.3} & \cellcolor{green!50}\textbf{25.2} & \cellcolor{green!50}\textbf{59.2} & \cellcolor{green!20}77.6\\ 
        \bottomrule
    \end{tabular}
    }
    \vspace{-6mm}
\end{table}

\section{Conclusion}
\label{sec:conclusion}
\vspace{-2mm}

We propose Hamba, a novel Mamba-based model for 3D hand reconstruction, which is capable of reconstructing robust 3D hand meshes with graph learning and state space modeling under bidirectional scanning. Our key insight is reformulating the Mamba scanning into graph-guided bidirectional scanning using a few effective tokens. This allows us to leverage the relations between hand joints and joint spatial sequences, addressing the occlusion and truncation problems using graph learning and state space modeling. Specifically, we designed a new GSS block to capture the relation between hand joints using graph convolution layers and Mamba blocks. Finally, we introduce a practical fusion module to boost performance by incorporating state space features and global features. Experiments on challenging benchmarks and in-the-wild tests demonstrate that Hamba outperforms all existing SOTA models.

\noindent \textbf{Limitations.} Although we leverage the strong representation capability from the graph-guided Mamba model and train on the large comprehensive datasets, it may still not be enough to cover all in-the-wild situations. Our current method lacks the capability to explore temporal features in videos because crawling video datasets requires extensive manual labor for 3D hand reconstruction.

\noindent \textbf{Broader Impacts.} Our research focuses on the Hamba model for 3D hand reconstruction, and we plan to release the pre-trained models and code. However, there is a potential risk that it could be used for unauthorized surveillance or privacy infringements.

\clearpage

\section*{Acknowledgments}

Aviral Chharia was supported in part by the ATK-Nick G. Vlahakis Graduate Fellowship from Carnegie Mellon University, USA. The authors would like to thank Bernhard Kerbl, Ce Zheng, Cheng Zhang, Yunlu Chen, and Zhenyu Xie for providing suggestions and feedback to improve this work.

\bibliographystyle{plain}
\bibliography{neurips_2024}

%%%%%%%%%%%%%%%%%%%%%%%%%%%%%%%%%%%%%%%%%%%%%%%%%%%%%%%%%%%%
\clearpage

\appendix
\renewcommand{\thefigure}{S\arabic{figure}}
\setcounter{figure}{0}
\renewcommand{\thetable}{S\arabic{table}}
\setcounter{table}{0}

\section{Appendix}
\label{Supp:Appendix}

We further elaborate on the additional model architecture details (Section~\ref{Supp: Detailed model structure}), and other training details~\ref{Supp:Model_training_and_other_details} including loss function weights, dataset descriptions, training schemes, test-time augmentation, and various evaluation metric definitions. We also include visual results for failure cases (Section~\ref{Supp:Failure Cases}). In Section~\ref{Supp:Additional_Details} we provide a detailed explanation of the ablation experiments, Table~\ref{tab:hint} and the algorithm of the GSS block. In Section~\ref{Supp:Ablation study with GCN + Transformer} we include an additional ablation to compare our proposed model with attention-based models. Finally 
Section~\ref{Supp:Transfer to hmr} demonstrates the transferability of our GSS block over the 3D Human Mesh Recovery task.

\subsection{Model Architecture Details}
\label{Supp: Detailed model structure}
\begin{figure}[h]
    \centering
    \includegraphics[width=\textwidth]{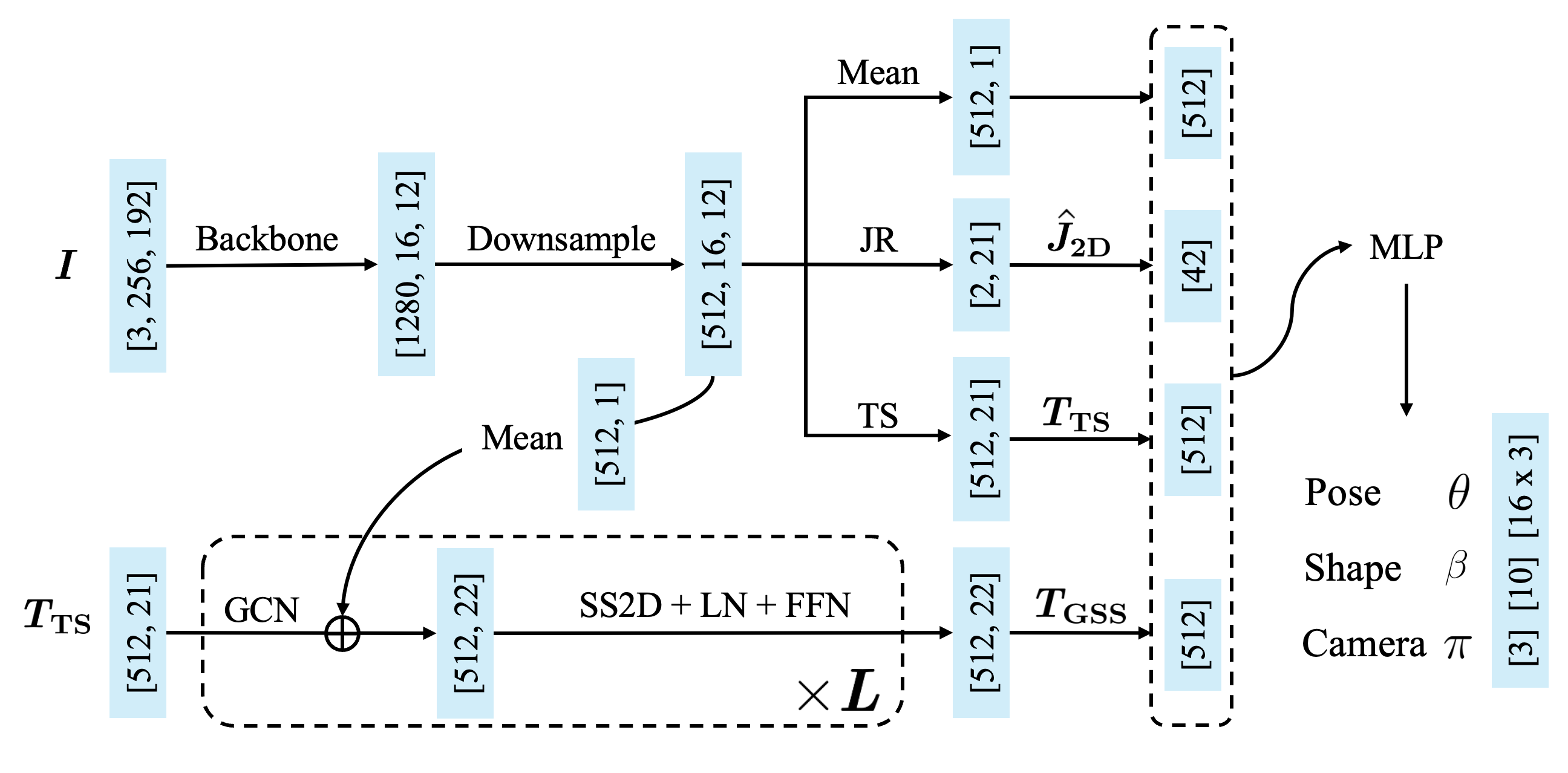}
    \vspace{-4mm}
    \caption{Illustration of model architecture feature dimensions for additional clarity.}
    \label{fig:feature_dimensions}
\end{figure}

Hamba follows an encoder-decoder structure that first tokenizes the input image patches, and then feeds it into the decoder to predict the 3D hand reconstruction results. Architecture details of each component required to reproduce our results are included in this section. We further include the model architecture feature dimensions in Figure~\ref{fig:feature_dimensions} for additional clarity.

\noindent\textbf{Backbone.} Following previous works~\cite{pavlakos2024reconstructing,vasu2023fastvit,zhou2024simple}, we ViT-H~\cite{dosovitskiy2020ViT} was used as the encoder that inputs an $I \in \mathbb{R}^{256\times 192\times 3}$ and tokenizes the image patches to output $T \in \mathbb{R}^{16 \times 12 \times 1280}$ tokens. Specifically, the backbone contained 50 transformer layers, with a total of 630M parameters. The downsampling layers consist of 2D convolution, Batch Norm, ReLU activation, and another 2D convolution. The channel dimension is reduced from 1280-dim to 512-dim in the first convolution operation, while it is maintained as 512-dim in the second. The kernel sizes for both convolutions were set as one.

\noindent\textbf{Joints Regressor (JR).} The JR follows a simple structure, containing only four VSSM (SS2D) blocks followed by 3 linear layers (MLP) to predict each of the MANO parameters. The VSSM blocks are 512-dim with a depth of 2, SSM state dimension of 1, SSM ratio of 2, MLP ratio of 4, and a depth-wise convolution kernel size of 3 (without bias). The Gate control is deprecated, and the JR takes all $T \in \mathbb{R}^{16 \times 12 \times 1280}$ tokens input. The final output of the VSSM blocks maintains the same shape as the input. `Mean' Pooling is performed over the height and width dimensions and then fed into the linear layers (MLP) to regress the MANO parameters $(\hat{\theta}, \hat{\beta}, \hat{\pi})$. 

\noindent \textbf{Token Sampler (TS).} The TS was implemented as the \texttt{pytorch.nn.Functional.grid\_sample} module of PyTorch. The TS is followed by a 1D Conv, Batch Norm, ReLU activation, and another 1D Conv operation. Bilinear Interpolation was used as the sampling mode to better adapt the float-point joint location prediction from the JR.

\noindent \textbf{Graph-guided State Space (GSS) Block.} The GSS Block has a simple structure. Note that each GCN layer in a GSS block consists of a single graph convolution operation~\cite{lee2018higherorder} followed by a Batch Norm and ReLU activation. For the rest of the components of the GSS blocks, each has a set dimension of 512-dim, a depth of 2, an SSM state dimension of 1, an SSM ratio of 2, an MLP ratio of 4, and a depth-wise convolution kernel size of 3 (without bias). The gate control is deprecated. We use 04 GSS blocks in our best-trained model.

\subsection{Model Training and other details}
\label{Supp:Model_training_and_other_details}

\noindent \textbf{Loss Functions.} As described by Equation~\ref{eq:loss_fn}, the L1 Norm is used to calculate the 3D joint loss and the 2D joint reprojection loss. Let $J^{\text{GT}}_{\text{2D}}$ and $J^{\text{GT}}_{\text{3D}}$ be the 2D and 3D ground-truth (GT) joint locations, while $J_{\text{2D}}$ and $J_{\text{3D}}$ be the corresponding model predictions. If a training dataset additionally provides the GT MANO parameters, i.e., $\theta^{\text{GT}}$ and $\beta^{\text{GT}}$ with their dataset, we also calculate a MANO parameter loss ($\mathcal{L}_{\theta}, \mathcal{L}_{\beta}$) between prediction ($\theta, \beta$) and the GT ($\theta^{\text{GT}}, \beta^{\text{GT}}$) using the L2 Norm. The loss terms are formulated as:\\
\vspace{-1em}
\begin{minipage}{0.5\textwidth}
\begin{equation}
\label{eq:3D_loss}
\begin{split}
  \mathcal{L}_{\text{3D}} &= ||J^{\text{GT}}_\text{3D}-J_\text{3D}||_1,  \\
  \mathcal{L}_{\text{2D}} &= ||J^{\text{GT}}_\text{2D}-J_\text{2D}||_1,
\end{split}
\end{equation}
\end{minipage}
\begin{minipage}{0.5\textwidth}
% \vspace{+1em}
\begin{equation}
\begin{split}
    \mathcal{L}_{\theta} &= ||\theta^{\text{GT}} - \theta||_2, \\
    \mathcal{L}_{ \beta} &= || \beta^{\text{GT}} - \beta||_2,
\end{split}\label{eq:2D_loss}
\end{equation}
\end{minipage}
\vspace{+1em}

To prevent the model from predicting unnatural hand gestures, we cooperate with discriminators $D_k$ for ($\theta,\beta$) and each hand joint angle separately following \cite{goel2023humans,kanazawa2018end,pavlakos2024reconstructing} as: $\mathcal{L}_{adv}= \sum_k(D_k(\theta,\beta)-1)^2$. The loss weights are kept as $\lambda_{\text{3D}}=0.05, \lambda_{\text{2D}}=0.01, \lambda_{\theta}=0.001, \lambda_{\beta}=0.0005$ and $\lambda_{\text{adv}}=0.0005$.

\noindent \textbf{Datasets.} \label{Supp:Datasets} In this section, we briefly introduce the datasets used in the study. For training, we use a mixture of FreiHAND~\cite{zimmermann2019freihand}, HO3D~\cite{hampali2020honnotate}, MTC~\cite{xiang2019monocular}, RHD~\cite{zimmermann2017learning}, Interhand2.6M~\cite{moon2020interhand2}, H2O3D~\cite{hampali2020honnotate}, DexYCB~\cite{chao2021dexycb}, COCO Wholebody~\cite{jin2020whole}, Halpe~\cite{fang2022alphapose}, and MPII NZSL~\cite{simon2017hand} datasets. The final dataset contained a total of 2.7M training samples. We adopt the same dataset mixing ratios as~\cite{pavlakos2024reconstructing} with sampling weights as 0.25 for FreiHAND and InterHand2.6M, and 0.1 for MTC and COCO-Wholebody. The rest were all set to 0.05.

\begin{itemize}
    \item \textbf{FreiHAND}~\cite{zimmermann2019freihand} is a large-scale multiview hand dataset with 3D hands annotations, popularly used by 3D hand reconstruction studies. The training set contains 33k samples collected by 08 cameras in a green-screen studio environment, which are further enhanced by replacing the backgrounds, leading to an overall 132k samples. The evaluation set contains 4K samples including both in-door and out-door in-the-wild scenarios. FreiHAND was released by Adobe Research and University of Freiburg in 2019, and is for research only, non-commercial use.

    \item \textbf{HO3Dv2}~\cite{hampali2020honnotate} and \textbf{HO3Dv3}~\cite{hampali2022keypoint} datasets are part of an ongoing international competition on 3D Hand Reconstruction. Both HO3Dv2 and HO3Dv3 are markerless hand-object interaction datasets containing 3D poses for both hand and object, released in 2020 and 2022 respectively. The sequences in the dataset came from the YCB dataset~\cite{xiang2017posecnn} and were collected by single or multiple RGBD cameras. Currently, the dataset has two versions: $v2$ and $v3$. The $v2$ version contains about 70k training images and around 10k test images, while the $v3$ version contains more than 103K training and 20K test images. The authors do not release the GT annotations and the results can only be tested using the Codalab~\cite{codalab_competitions_JMLR} competition website. Hamba achieves the best performance (as of May 2024) on the leaderboard of both datasets. HO3D was released by Graz University of Technology and CNRS France, and is for research only, non-commercial use.

    \item \textbf{HInt-EpicKitchensVISOR}~\cite{VISOR2022,pavlakos2024reconstructing}, \textbf{HInt-NewDays}~\cite{pavlakos2024reconstructing} and \textbf{HInt-Ego4D}~\cite{grauman2022ego4d} datasets consists of 40.4k hands with 2D keypoints, and was released in 2024. HInt stands for Hand Interactions in the Wild, which only has 2D labels and visibility labels. The three subsets 
 of HInt are built based on the Hands23~\cite{cheng2023towards}, Epic-Kitchens~\cite{damen2018scaling}, and the Ego4D~\cite{grauman2022ego4d}  video datasets. In our study, HInt is not used for training but serves as a benchmark to evaluate Hamba's cross-dataset generalizability. This dataset is for research only and non-commercial use.
\end{itemize}

In addition to the above-discussed datasets, we include the descriptions of the datasets used in the Hamba training set:
\begin{itemize}
    \item \textbf{H2O3D}~\cite{hampali2022keypoint} is first released in 2022. Similar to HO3D, it contains around 61k training images and 9k test images using a five RGBD camera multi-view setup in a controlled environment. Compared to HO3D, the content is further expanded to two hands interacting with different objects. 

    \item \textbf{DexYCB}~\cite{chao2021dexycb} is a hand-manipulating object dataset captured by 8 RGBD cameras in a laboratory environment. It contains 582k RGBD images for 10 subjects grasping 20 different objects. Each of the sequences lasts for 3 seconds. The dataset comes with the 2D and 3D annotation for hand.

    \item \textbf{MTC}~\cite{xiang2019monocular} used the Panoptic Studio to capture both the 3D body and hand poses without using markers. The dataset recorded 40 subjects for 2.5 minutes and included 111K hand images and their 3D pose. Each of the subjects performed a large range of both body and hand motions.
    
    \item \textbf{RHD}~\cite{zimmermann2017learning} proposed a synthetic dataset for better annotation quality. The dataset utilized 20 different 3D characters performing 39 actions, containing around 44k images in total. Each image is collected under $320\times320$ resolution and comes with 3D joint annotation. To enhance the generalization of the dataset, the rendered hands are put on random background images. 
    
    \item \textbf{InterHand2.6M}~\cite{Moon_2020_ECCV_InterHand2.6M} is a large-scale real-captured hand interaction dataset. It contains 2.6M labeled RGB images showing single or two interacting hands. The data was collected using more than 80 cameras in a precisely calibrated multi-view studio and used a semi-automatic annotation method. The dataset involved 27 subjects and the image resolution is set to $512\times334$.
    
    \item \textbf{COCO Wholebody}~\cite{jin2020whole} based on the COCO dataset and annotated 133 whole body keypoints including face, hand, body, and feet. It contains 200k RGB in-the-wild images, with 250k instances, and also comes with face, hand, and body bounding boxes. As for the hand pose estimation task, it includes about 100k samples with 2D keypoint annotations.
    
    \item \textbf{Halpe}~\cite{fang2022alphapose} is an in-the-wild RGB dataset for full-body human-object interaction dataset. It contains 50k training and 5k test instances. Images are annotated with 136 full body keypoints.
    
    \item \textbf{MPII NZSL}\cite{simon2017hand} is a mixing dataset, containing in-the-wild, multiview studio collected and synthetic RGB images. Images are included with single-hand, hand-hand, and hand-object interactions. The dataset has 11k rendered synthetic samples, 2.8k manually annotated in-the-wild images from MPII~\cite{andriluka20142d} and NZSL~\cite{mckee2013making}, and 15k multiview samples from Panoptic Studio dataset~\cite{joo2015panoptic}. All the samples come with 2D keypoint annotations.
\end{itemize}

\noindent \textbf{Training schemes.} The Joints Regressor (JR) was trained on a single NVIDIA A4500 GPU with a batch size of 8 for 1M steps. The training took 5 days and required around 300GB RAM. The complete Hamba model was trained on two NVIDIA A6000 GPUs which required two days on a batch size of 56. Early stopping was used after 170k steps to prevent overfitting. The mixing ratio was kept consistent for both parts of the model. \texttt{AdamW} optimizer was used with a learning rate of \texttt{1e-5}, $\beta_1=0.9, \beta_2=0.999$, and a weight decay of \texttt{1e-4}. The ViT backbone with initialized with HaMeR~\cite{pavlakos2024reconstructing} released checkpoint and is kept unfroze during training. The same setting was used while training the JR. Note that for reproducing the results, we recommend directly loading the released Hamba weights from the GitHub repository. For training from scratch, we recommend using the same training configuration, since PyTorch Lightening might have a bug for effective batch size. Refer to the GitHub issue~\footnote{DDP with two GPUs doesn't give the same results as one GPU with the same effective batch size \#6789 \url{https://github.com/Lightning-AI/pytorch-lightning/issues/6789})}.

\noindent \textbf{Test Time Augmentation (TTA).} Since the FreiHAND dataset was originally part of a challenge on 3D Hand Mesh Reconstruction, many popular models used Test-time augmentation (TTA) to report their results. For a fair comparison with the models that used TTA, we report both the results, with and without TTA on the FreiHAND test set in Table~\ref{tab:freihand_results}. Following~\cite{lin2021mesh}, we let the model infer on the same test image, which was scaled and rotated multiple times and averaged over the results. This generally results in a performance boost. The rotations are set from -90\textdegree \ to 90\textdegree \ for every 10\textdegree, and rescaled by factors of (0.7, 0.8, 0.9, 1.0, 1.1). We tested through all the combinations and averaged the error to obtain the TTA result. Note that we obtained the SOTA results both with and without TTA. Unless stated otherwise, all reported results in our study are without using TTA.

\noindent \textbf{Evaluation Metrics.} Here, we discuss the definitions of the evaluation metrics used in the study.
\begin{itemize}
    \item \textbf{PA-MPJPE} means Procrustes Aligned-Mean Per Joint Position Error, which is the measure of the average joint error after Procrustes alignment. It is calculated as the Euclidean distance between the predicted joints and the GT. PA-MPJPE performs the Procrustes Analysis (PA)~\cite{gower1975generalized} method that aligns the predicted and GT positions using a non-rigid transformation to minimize the overall distance between them. The PA-MPJPE is measured in millimeters (mm) and widely used as the evaluation metrics in 3D hand reconstruction works.
    
    \item \textbf{PA-MPVPE} means Procrustes Aligned-Mean Per Vertex Position Error, which is similar to PA-MPJPE but measures error between mesh vertices after Procrustes alignment. It is also calculated in millimeters (mm).
    
    \item \textbf{F@5/ F@15mm} is the harmonic mean between recall and precision under a specified distance threshold. F@5 and F@15 represent the F-score with a threshold of 5mm and 15mm respectively.

    \item \textbf{PCK} stands for Percentage of Correct Keypoints. To find PCK, first, the Euclidean distances between the predicted and GT keypoint are normalized by head segment length. When the difference is under a certain threshold, the keypoint is considered to be as correct. By varying the threshold we can draw the curve for PCK that increases along with the threshold. 

    \item \textbf{AUC} denotes the Area Under the Curve. In line with previous works~\cite{lin2021mesh}, we specify AUC for the PCK curve with error thresholds between 0mm and 50mm. AUC for joints and mesh vertices are denoted as $\text{AUC}_J$ and $\text{AUC}_V$ respectively.
\end{itemize}

\subsection{Failure Cases}
\label{Supp:Failure Cases}
\begin{figure}
    \centering
    \includegraphics[width=\textwidth]{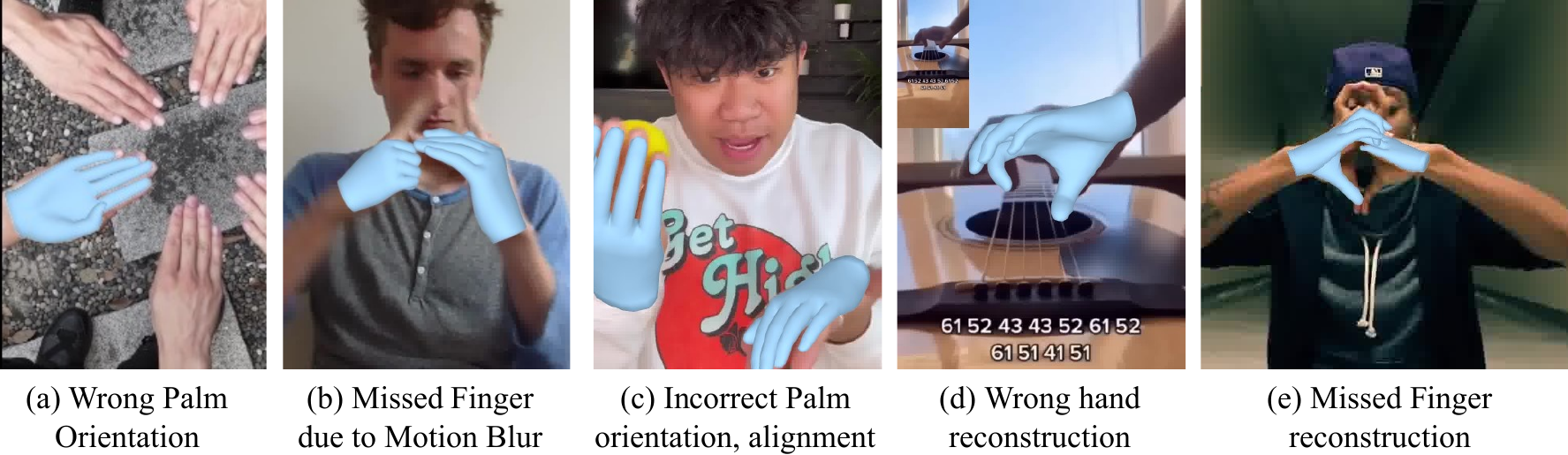}
    \vspace{-4mm}
    \caption{Illustration of various failure cases of Hamba: Wrong palm orientation, missed frames due to motion blur, and missed finger reconstruction.}
    \label{fig:failure_samples}
    \vspace{-3mm}
\end{figure}

 Figure~\ref{fig:failure_samples} presents the various failure cases encountered by Hamba during difficult in-the-wild scenarios. Here, we see that most failure cases occur when the model fails to robustly predict the palm orientation as shown in Figure~\ref{fig:failure_samples}(a, c); or when it misses the finger due to motion blur in videos captured at low FPS like Figure~\ref{fig:failure_samples}(b). Sometimes, the model cannot distinguish the direction where the palm is facing, thus making the prediction rotate for 180. Other cases include failure due to the detector predicting the wrong hand or complex finger gestures, as illustrated in Figure~\ref{fig:failure_samples}(e).

\subsection{Additional details}
\label{Supp:Additional_Details}

\noindent\textbf{Algorithm of GSS block.} To provide a more detailed description of the proposed Graph-guided State Space (GSS) block, we present the processing steps of the algorithm, as illustrated in Algorithm~\ref{algo:gss}.

\noindent\textbf{Additional details of Table~\ref{tab:hint}.} Since HInt~\cite{pavlakos2024reconstructing} provides annotations for occluded 2D hand keypoints, we report separate results considering: (i) all joints, (ii) only the joints annotated as visible, and lastly, (iii) only considering joints annotated as occluded in Table~\ref{tab:hint}. We notice that visible joints have a relatively higher PCK compared to all joints which include both the visible and occluded joints; while the occluded joints reconstruction has the lowest PCK. It is important to note that since 3D GT poses cannot be annotated for in-the-wild images, HInt only provides 2D keypoints. Therefore, we reproject the 3D hand mesh back to 2D and use PCK to evaluate the reprojection accuracy.

\noindent\textbf{Ablation Experiments.} This section provides a detailed explanation of the network architecture modifications made for the ablation experiments. The ablation results discussed below are presented in Rows 1-9 of Table~\ref{tab:ablation}. The first four experiments involve branch-wise ablations, while the remaining five focus on architectural modifications in the component-wise ablations. Including branch-wise experiments helps investigate the contribution at the branch level, while component-wise ablations clarify the improvements brought by the proposed component.
\begin{itemize}
    \item \textbf{w/o Token\_Sampler\_Branch.} We remove the TS branch's tokens from concatenation with other tokens in the fusion module, but we do not remove the token sampler. Thus the sampled tokens from the TS are still input to the GSS block.
    
    \item \textbf{w/o 2D\_Joints\_Feature\_Branch.} We removed the 2D joints feature from the JR, which is concatenated with the other tokens in the fusion module. The joint output from the JR is still used as a strong local context for sampling effective tokens.

    \item \textbf{w/o GSS\_Token\_Branch.} To evaluate the contribution of the GSS branch tokens, we remove the GSS tokens from the fusion module. This helps to evaluate the overall contribution of the GSS block tokens. It is important to note that modeling joint spatial relations provides better tokens than directly using the 2D joint locations. It is confirmed by the larger performance drop observed when the GSS tokens are removed compared to TS tokens, and 2D joints features.

    \item \textbf{w/o Global\_Mean\_Token\_Branch.} We remove the Global Mean Token branch from the model architecture and do not include their tokens in the fusion module. Note that we do not remove the global mean token, which is concatenated with the output of the GCN in the GSS block.
    
    \item \textbf{w/o Token\_Sampler.} We exclude the TS component directly.
    
    \item \textbf{w/o Bidirectional\_Scan.} To verify the effectiveness of the bidirectional scanning, we replaced it with unidirectional scanning. 
    
    \item \textbf{w/o GCN.} We remove the GCN module to verify the effectiveness of GCN. The sampled tokens are directly concatenated with the Global Mean Token and input to the SS2D Block.
    
    \item \textbf{w/o Graph-guided\_Bi\_scan.} We shuffled the order of the joint sequence to simulate without graph-guided scanning.
    
    \item \textbf{w/o Mamba (SS2D+LN+FFN).} To evaluate the contribution of Mamba blocks (SS2D+LN+FFN) in the GSS. We remove the Mamba blocks from the proposed GSS blocks. This includes the SS2D, Layer Norm (LN), and the Feed-Forward Network (FFN). The GSS blocks degenerate into simple GCN blocks. We still concatenate it with the global mean token with the output of the GSS block and split it before feeding it into the next GSS block.
\end{itemize}

\algsetup{linenosize=\footnotesize}
\vspace{-4mm}
\begin{center}
\begin{algorithm}[H]
    \caption{Graph-guided State Space (GSS) block} 
    \label{algo:gss}
    \footnotesize
    \renewcommand{\algorithmicrequire}{\textbf{Input:}}
    \renewcommand{\algorithmicensure}{\textbf{Output:}}
    \setlength\medmuskip{0mu}
    \begin{algorithmic}[1]
        \REQUIRE $\text{Feature~Representation} ~T: ~\textcolor{green!80!black}{(B, C, H, W)}$
        \ENSURE $\text{Enhanced~Representation} ~T_{\text{GSS}_{L}}: ~\textcolor{green!80!black}{(B, C, J)}$
        \STATE \textcolor{red!80!black}{/* ~Token ~Sample~*/}
        \STATE $T_{\text{TS}}: \textcolor{green!80!black}{(B,C,J')} \leftarrow TS(Conv2D(T), \textcolor{gray!60}{\hat{J}_{2D}})$
        \STATE $m: \textcolor{green!80!black}{(B,C,1)} \leftarrow Mean(T)$
        \STATE \textcolor{red!80!black}{/* ~A ~Set ~of ~GSS ~blocks~*/}
        \STATE $T_{\text{GSS}_l}^{\{1, .., 21\}} \leftarrow T_{\text{TS}}$ when $l=0$
        \FOR{$l$ in [1 to L]}
            \STATE \textcolor{red!80!black}{/* ~Graph ~Convolutions ~for ~Hand ~Joints*/}
            \STATE $T_{\text{GCN}_l} : \textcolor{green!80!black}{(B, J', C)}\leftarrow GCN(T_{\text{GSS}_{l-1}}^{\{1, .., 21\}})$
            \STATE $z' : \textcolor{green!80!black}{(B, C, J)}\leftarrow Concat(T_{\text{GCN}_l}, m)$
            \STATE \textcolor{red!80!black}{/* ~SS2D ~Block */}
            \STATE $z'' : \textcolor{green!80!black}{(B, C, J)}\leftarrow Linear(Norm(z'))$
            \STATE $z''' : \textcolor{green!80!black}{(B, C, J)}\leftarrow SiLU(DWConv(z''))$
            \STATE $z_{\text{SS2D}} : \textcolor{green!80!black}{(B, C, J)}\leftarrow SS2D(z''')$
            \STATE $z_{\text{SS2D}}': \textcolor{green!80!black}{(B, C, J)}\leftarrow Linear(LayerNorm(z_{\text{SS2D}}))$
            \STATE \textcolor{red!80!black}{/* ~Residual ~Connection */}
            \STATE $y' : \textcolor{green!80!black}{(B, C, J)}\leftarrow Sum(z', z_{\text{SS2D}}')$
            \STATE \textcolor{red!80!black}{/* ~FFN ~after ~Layer ~Normalization */}
            \STATE $y'' : \textcolor{green!80!black}{(B, C, J)}\leftarrow FFN(LayerNorm(y'))$
            \STATE \textcolor{red!80!black}{/* ~Residual ~Connection */}
            \STATE $T_{\text{GSS}_l} : \textcolor{green!80!black}{(B, C, J)}\leftarrow Sum(y', y'')$
            \STATE $T_{\text{GSS}_{l-1}}^{\{1, .., 21\}}, m  \leftarrow Split(T_{\text{GSS}_l})$
        \ENDFOR
        \STATE \textcolor{red!80!black}{/* ~Enhance ~feature ~representation ~*/}
        \STATE $T_{\text{GSS}_L} : \textcolor{green!80!black}{(B, C, J)} \leftarrow Concat(T_{\text{GSS}_{L}}^{\{1, .., 21\}}, m)$
        \RETURN $T_{\text{GSS}_L}$
    \end{algorithmic}
\end{algorithm}
\end{center}
\vspace{-4mm}

\begin{figure}[t]
    \centering
    \includegraphics[width=\textwidth]{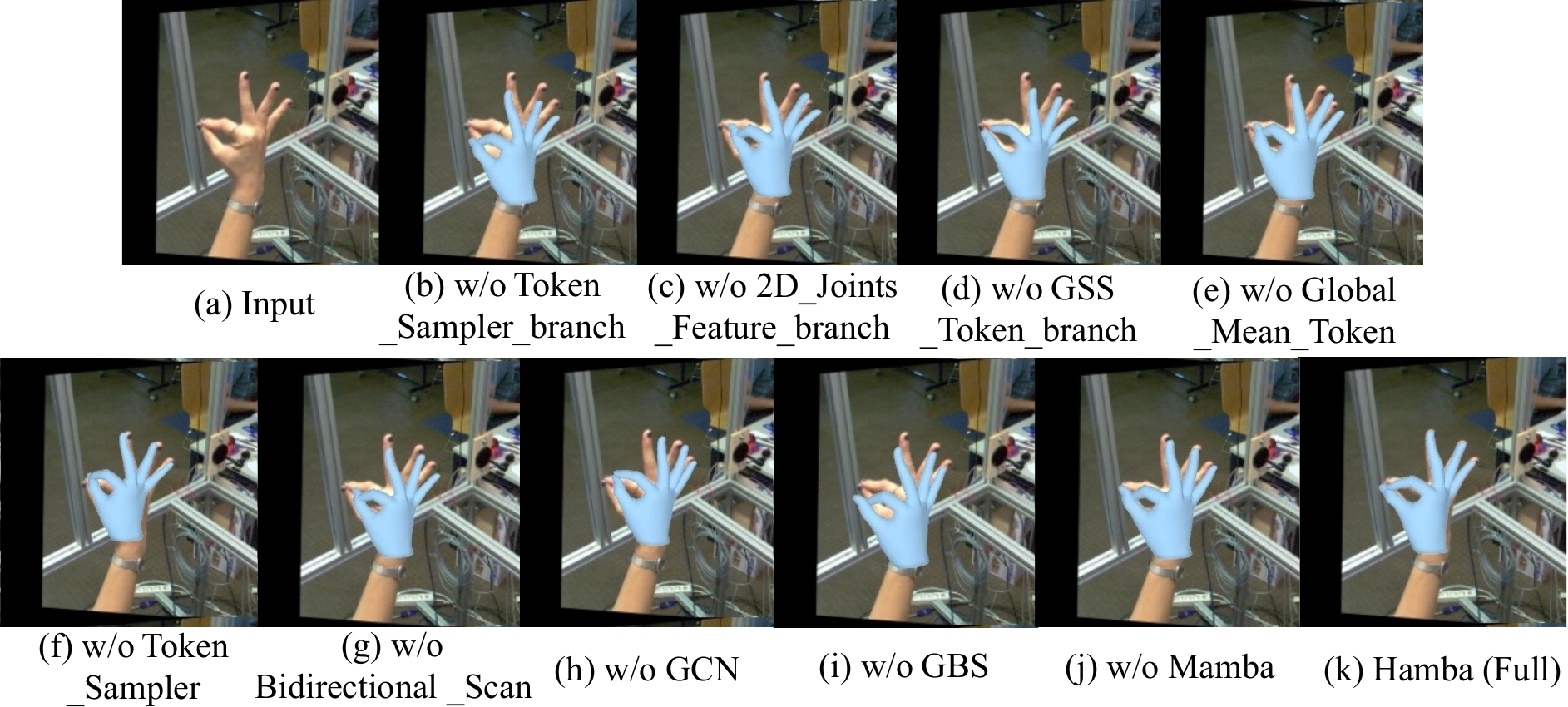}
    \vspace{-4mm}
    \caption{\textbf{Qualitative Ablation Study} on FreiHAND~\cite{zimmermann2019freihand}. Our full model achieves the best result compared with all the ablation variants.}
    \label{fig:ablation_visualization}
    \vspace{-4mm}
\end{figure}

\subsection{Ablation study with GCN + Transformer models}
\label{Supp:Ablation study with GCN + Transformer}
We performed an additional ablation study to compare the effect of SS2D block with attention-based method. The GCN + SS2D in Graph-guided state space block is replaced with GCN + Attention borrowed from Graformer~\cite{zhao2022graformer}. The comparison results are shown in Table~\ref{tab:additional_ablation_results}. Both models are trained with the same dataset setting on a single A6000 GPU for 60K steps, and evaluated on the FreiHand dataset~\cite{zimmermann2019freihand}. Our proposed GCN + SS2D model shows improvement in all metrics compared with the GCN + Transformer architecture. This confirms our graph-guided state-space model has better capability to learn the hand joint relationships.

\begin{table}[H]
\renewcommand{\arraystretch}{0.80}
\setlength{\tabcolsep}{3pt}
\caption{Ablation study on the FreiHand dataset~\cite{zimmermann2019freihand} to verify the effectiveness of GCN + SS2D block comparing with GCN + attention mechanism. Here, `w' denotes `with'.}
  \label{tab:additional_ablation_results}
  \centering
  \resizebox{0.7\linewidth}{!}{
    \begin{tabular}{l|cccc}
    \toprule
    Method & PA-MPJPE$~\downarrow$ & PA-MPVPE$~\downarrow$ & F@5mm$~\uparrow$ & F@15mm$~\uparrow$\\
    \midrule
    w GCN + Attention & 7.0 & 6.6 & 0.730 & 0.985\\
    w GCN + SS2D (Ours) & \bf 6.6 & \bf 6.3 & \bf 0.738 & \bf 0.988\\
    \bottomrule
    \end{tabular}
}
\end{table}

Compared to transformer-based models that utilize a large number of tokens for 3D hand reconstruction, our proposed Hamba uses fewer tokens and is `token-efficient'. We provide the details of Hamba method's efficiency in terms of inference time, FLOPs, and GPU memory usage in Table~\ref{tab:efficiency_results}. This shows our model is more lightweight and faster comparing to GCN + Transformer-based models.

\begin{table}[H]
\renewcommand{\arraystretch}{0.80}
\setlength{\tabcolsep}{3pt}
\caption{Comparison of Token Efficiency, Parameters, FLOPS, Runtime and GPU Memory.}
  \label{tab:efficiency_results}
  \vspace{-1mm}
  \centering
  \resizebox{1\linewidth}{!}{
    \begin{tabular}{lc|cccc|c|ccc|c}
    \toprule
    \multirow{2}{*}{Method} & \multirow{2}{*}{Tokens$~\downarrow$} & \multicolumn{4}{c|}{Parameters (M)$~\downarrow$} & MFLOPs$~\downarrow$ & \multicolumn{3}{c|}{Runtime (ms)$~\downarrow$} & GPU$~\downarrow$ \\
    &  & Backbone & JR & Decoder & All & Decoder & Backbone & JR & Decoder & Mem. (MB) \\
    \midrule
    w GCN + Transformer & 192 & 630 & 27.6 & 149 & 782 & 830 & 18.7 & 9.0 & 21.9 & 20947 \\
    w GCN + SS2D (Ours) & 22 & 630 & 27.6 & 71.8 & 733 & 649 & 18.7 & 9.0 & 11.8 & 3413.2 \\
    \midrule
    & \cellcolor{green!50} \bf 88.5\%$\downarrow$ & - & - & \cellcolor{green!50} \bf 51.8\%$\downarrow$ & \cellcolor{green!50} \bf 6\%$\downarrow$ & \cellcolor{green!50} \bf 21.8\%$\downarrow$ & - & - & \cellcolor{green!50} \bf 46.1\%$\downarrow$ & \cellcolor{green!50} \bf 83.7\%$\downarrow$ \\
    \bottomrule
    \end{tabular}
}
    \vspace{-2mm}
\end{table}

\subsection{Transfer to 3D Human Mesh Recovery task}
\label{Supp:Transfer to hmr}
\begin{figure}[H]
  \centering
  \includegraphics[width=\textwidth]{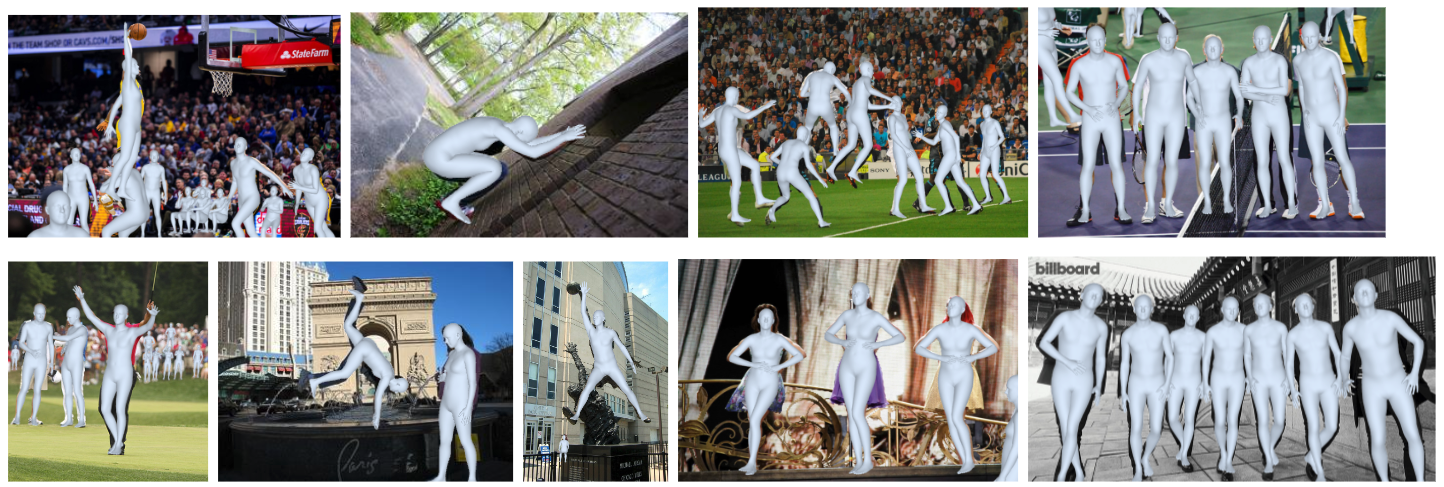}
  \vspace{-6mm}
  \caption{Visual Results of Hamba for Full body Human Reconstruction}
  \label{fig:full_body}
\end{figure}

To test the transferability of the GSS block acting as a plug-and-play module for other downstream tasks, we adapted Hamba for the 3D human mesh recovery (HMR) task. We trained our model on the same mixing datasets as 4D-humans~\cite{goel2023humans}. Our model achieved comparable performance with 4D-humans (HMR2.0b) as shown in Table~\ref{tab:transfer_results}. Hamba showed improvements on LSP-Extended~\cite{LSP-Extended} and COCO datasets~\cite{lin2014coco}, as well as achieving comparable results on the 3DPW dataset~\cite{von2018recovering3dpw}, even though it's trained for fewer steps on a single GPU. The performance of our model may be further improved by increasing training iterations as HMR2.0b~\cite{goel2023humans} under 8 GPU settings. This confirms our proposed module is capable of serving as a plug-and-play component to solve similar or downstream tasks. The visual results for in-the-wild scenarios are shown in Figure~\ref{fig:full_body}.

\begin{table}[H]
\renewcommand{\arraystretch}{0.80}
\caption{Results comparison on the Human Mesh Recovery task on three Benchmark datasets.}
  \label{tab:transfer_results}
  \vspace{-1mm}
  \centering
  \resizebox{1\linewidth}{!}{
    \begin{tabular}{lc|cc|cc|cc}
    \toprule
    \multirow{2}{*}{Method} & \multirow{2}{*}{Training Settings} & \multicolumn{2}{c|}{LSP-Ext} & \multicolumn{2}{c|}{COCO} & \multicolumn{2}{c}{3DPW} \\
    &   & @0.05 $\uparrow$ & @0.1$\uparrow$ & @0.05$\uparrow$ & @0.1 $\uparrow$ & MPJPE$\downarrow$ & PA-MPJPE$\downarrow$\\
    \midrule
    HMR2.0b~\cite{goel2023humans} & 8 $\times$ A100s, 1M Steps & 0.530 & 0.820 & \cellcolor{green!50} \bf 0.860 & 0.960 & \cellcolor{green!50} \bf 81.3 & \cellcolor{green!50} \bf 54.3 \\
    Hamba (Ours)	& 1 $\times$ A100, 300K Steps & \cellcolor{green!50} \bf 0.539 & \cellcolor{green!50} \bf 0.832 & 0.856 & \cellcolor{green!50} \bf 0.966 & 81.7 & 54.7 \\
    \bottomrule
    \end{tabular}
}
    \vspace{-2mm}
\end{table}

\clearpage

\begin{figure}[H]
  \centering
  \includegraphics[width=\textwidth]{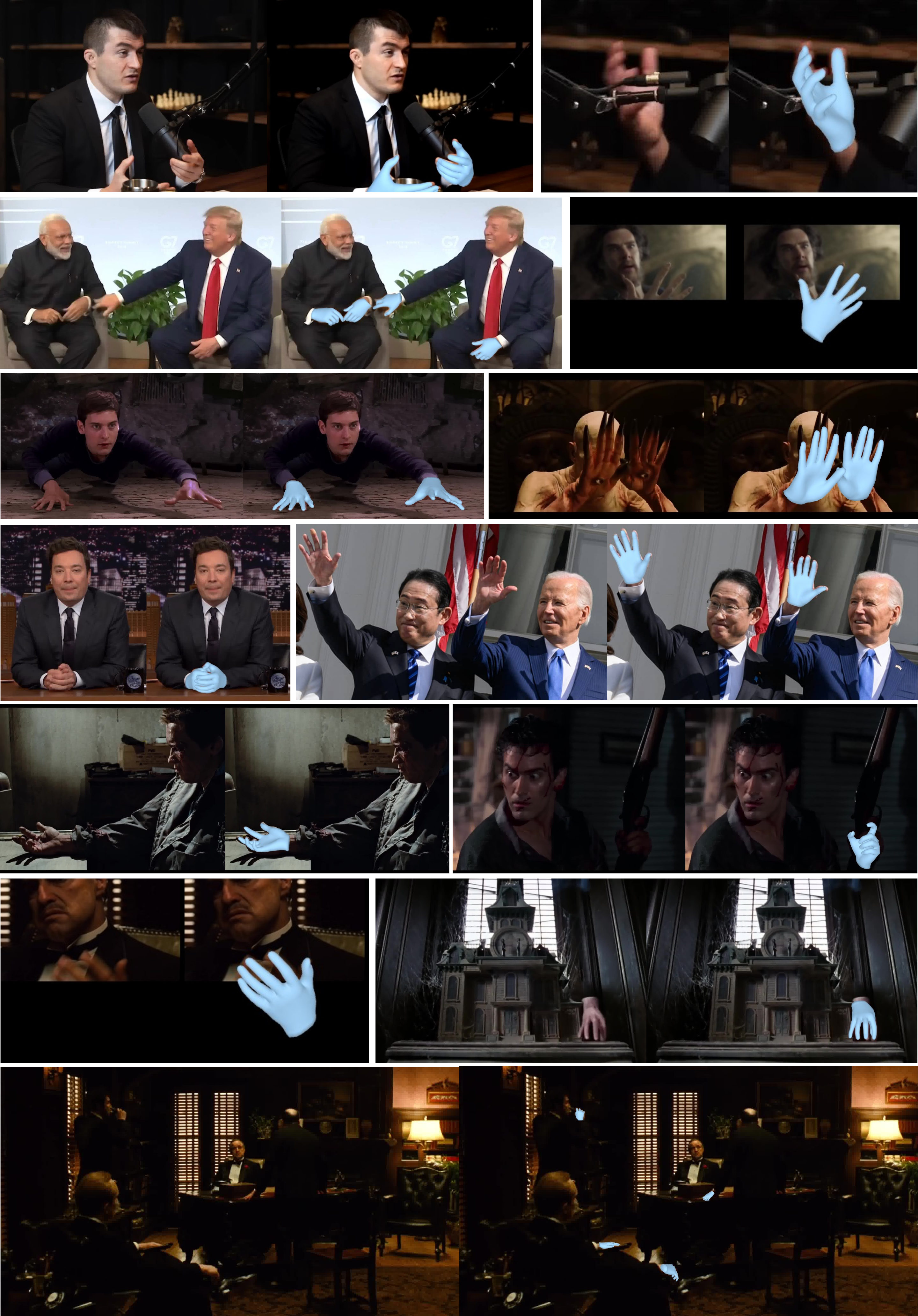}
  \vspace{-2mm}
  \caption{\textbf{Additional in-the-wild visual results of Hamba}. Hamba achieves significant performance in various in-the-wild scenarios, including truncations, hands interacting with objects or hands, different skin tones, viewpoints, angles, occlusion, and movie scenes.}
  \label{fig:additional_results}
  \vspace{-4mm}
\end{figure}

\begin{figure}[H]
  \centering
  \includegraphics[width=\textwidth]{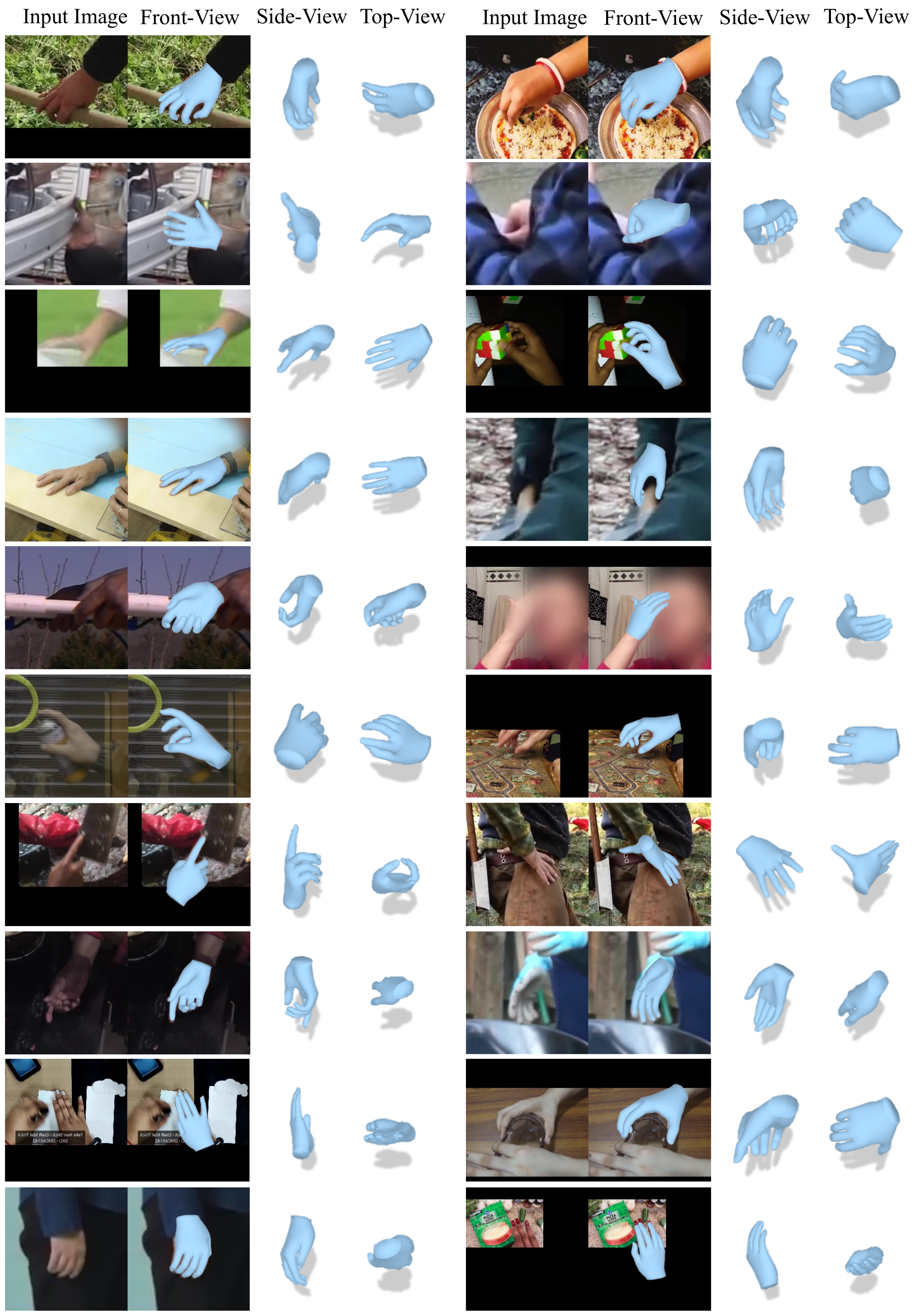}
  \vspace{-1em}
  \caption{\textbf{Qualitative Results on HInt-NewDays}~\cite{cheng2023towards,pavlakos2024reconstructing}. In-the-wild testing results of Hamba on the HInt-NewDays, which includes highly-occluded hands, hand-hand or hand-object interactions, and truncation scenarios. We did not use the HInt dataset to train Hamba.}
  \label{fig:new_days_in_the_wild}
  \vspace{-0.5em}
\end{figure}

\begin{figure}[H]
  \centering
  \includegraphics[width=\textwidth]{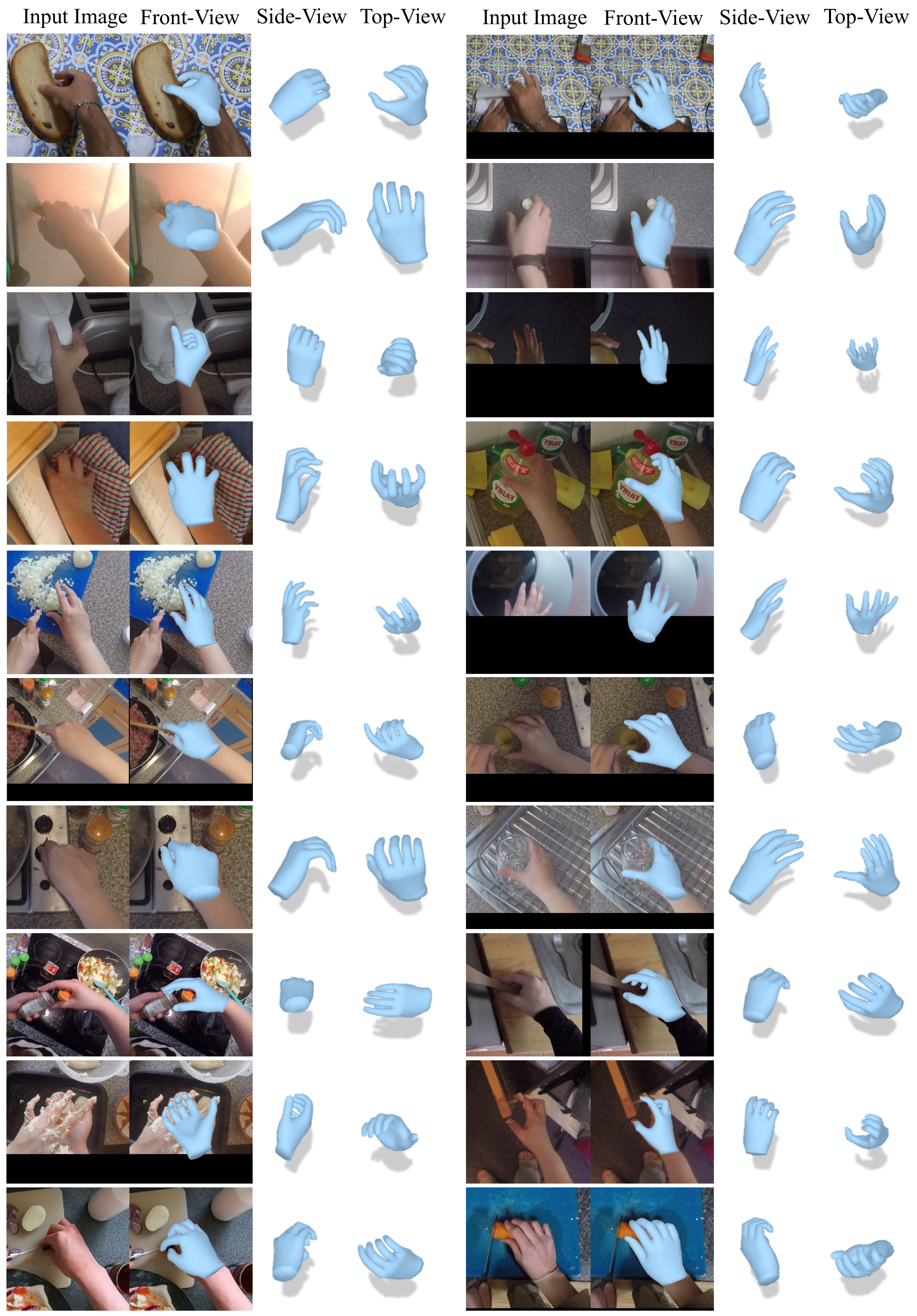}
  \vspace{-1em}
  \caption{\textbf{Qualitative Results on HInt-EpicKitchensVISOR}~\cite{damen2018scaling,pavlakos2024reconstructing}. In-the-wild testing results of Hamba on the HInt-EpicKitchensVISOR, which includes challenging cooking videos. We did not use the HInt dataset to train Hamba.}
  \label{fig:VISOR_in_the_wild}
  \vspace{-0.5em}
\end{figure}

%%%%%%%%%%%%%%%%%%%%%%%%%%%%%%%%%%%%%%%%%%%%%%%%%%%%%%%%%%%%

\newpage
\section*{NeurIPS Paper Checklist}

\begin{enumerate}

\item {\bf Claims}
    \item[] Question: Do the main claims made in the abstract and introduction accurately reflect the paper's contributions and scope?
    \item[] Answer: \answerYes{} % Replace by \answerYes{}, \answerNo{}, or \answerNA{}.
    \item[] Justification: {\color{blue} The claims made in the abstract and the contribution section of the Introduction are widely substantiated with 9 ablation studies, including visual comparisons for ablations (see Section~\ref{sec:ablation}), as well as experiments on 4 popular benchmark datasets.}

\item {\bf Limitations}
    \item[] Question: Does the paper discuss the limitations of the work performed by the authors?
    \item[] Answer: \answerYes{} % Replace by \answerYes{}, \answerNo{}, or \answerNA{}.
    \item[] Justification: {\color{blue} The limitations are discussed in Section~\ref{sec:conclusion}. Although we leverage the strong representation capability from the graph-guided Mamba model and train on the large comprehensive datasets, our experiments do not fully explore temporal features from videos due to the high cost of collecting video datasets for 3D hand reconstruction.}

\item {\bf Theory Assumptions and Proofs}
    \item[] Question: For each theoretical result, does the paper provide the full set of assumptions and a complete (and correct) proof?
    \item[] Answer: \answerNA{} % Replace by \answerYes{}, \answerNo{}, or \answerNA{}.
    \item[] Justification: {\color{gray} We do not propose a theoretical proof. This question is not applicable to our work.}
    \item[] Guidelines:
    \begin{itemize}
        \item The answer NA means that the paper does not include theoretical results. 
        \item All the theorems, formulas, and proofs in the paper should be numbered and cross-referenced.
        \item All assumptions should be clearly stated or referenced in the statement of any theorems.
        \item The proofs can either appear in the main paper or the supplemental material, but if they appear in the supplemental material, the authors are encouraged to provide a short proof sketch to provide intuition. 
        \item Inversely, any informal proof provided in the core of the paper should be complemented by formal proofs provided in the appendix or supplemental material.
        \item Theorems and Lemmas that the proof relies upon should be properly referenced. 
    \end{itemize}

    \item {\bf Experimental Result Reproducibility}
    \item[] Question: Does the paper fully disclose all the information needed to reproduce the main experimental results of the paper to the extent that it affects the main claims and/or conclusions of the paper (regardless of whether the code and data are provided or not)?
    \item[] Answer: \answerYes{} % Replace by \answerYes{}, \answerNo{}, or \answerNA{}.
    \item[] Justification: {\color{blue} We have described our model architecture in detail in Appendix Section~\ref{Supp: Detailed model structure}. The training schemes have been included in Section~\ref{Supp:Model_training_and_other_details}.}

\item {\bf Open access to data and code}
    \item[] Question: Does the paper provide open access to the data and code, with sufficient instructions to faithfully reproduce the main experimental results, as described in supplemental material?
    \item[] Answer: \answerYes{}{} % Replace by \answerYes{}, \answerNo{}, or \answerNA{}.
    \item[] Justification: {\color{blue} Our code was included in the Supplementary .zip file during the NeurIPS review. We will open-source it shortly with a detailed readme on the project's Github repository.}

\item {\bf Experimental Setting/Details}
    \item[] Question: Does the paper specify all the training and test details (e.g., data splits, hyperparameters, how they were chosen, type of optimizer, etc.) necessary to understand the results?
    \item[] Answer: \answerYes{} % Replace by \answerYes{}, \answerNo{}, or \answerNA{}.
    \item[] Justification: {\color{blue} The experimental settings including the data splits, hyperparameters, optimizers, etc., are included in the Experiments section of the manuscript under Experimental settings. Additional model settings and training details are included in the Appendix section Section~\ref{Supp: Detailed model structure} and Section~\ref{Supp:Model_training_and_other_details}.}

\item {\bf Experiment Statistical Significance}
    \item[] Question: Does the paper report error bars suitably and correctly defined or other appropriate information about the statistical significance of the experiments?
    \item[] Answer: \answerNo{} % Replace by \answerYes{}, \answerNo{}, or \answerNA{}.
    \item[] Justification: {\color{orange} We test on large datasets. Some of them restrict access to test sets and we are only able to evaluate through their website competition, which will not report the error for each sample. This makes it impractical even impossible to conduct the error analysis. The same methodology was followed by the previous works on the same topic.}

\item {\bf Experiments Compute Resources}
    \item[] Question: For each experiment, does the paper provide sufficient information on the computer resources (type of compute workers, memory, time of execution) needed to reproduce the experiments?
    \item[] Answer: \answerYes{} % Replace by \answerYes{}, \answerNo{}, or \answerNA{}.
    \item[] Justification:{\color{blue} The type of GPU, RAM, memory, compute workers, and other compute-related parameters have been included in the Supplementary Section~\ref{Supp:Model_training_and_other_details}.}
    
\item {\bf Code Of Ethics}
    \item[] Question: Does the research conducted in the paper conform, in every respect, with the NeurIPS Code of Ethics \url{https://neurips.cc/public/EthicsGuidelines}?
    \item[] Answer: \answerYes{} % Replace by \answerYes{}, \answerNo{}, or \answerNA{}.
    \item[] Justification: {\color{blue} The research conducted in the paper conforms, in every respect, with the NeurIPS Code of Ethics.}

\item {\bf Broader Impacts}
    \item[] Question: Does the paper discuss both potential positive societal impacts and negative societal impacts of the work performed?
    \item[] Answer: \answerYes{}{} % Replace by \answerYes{}, \answerNo{}, or \answerNA{}.
    \item[] Justification: {\color{blue} We have included both the potential positive and negative social impacts of the work along with the limitations. We will release the pre-trained models and source code. It may be used for unwarranted surveillance or privacy violations.}
    
\item {\bf Safeguards}
    \item[] Question: Does the paper describe safeguards that have been put in place for responsible release of data or models that have a high risk for misuse (e.g., pretrained language models, image generators, or scraped datasets)?
    \item[] Answer: \answerNA{}{} % Replace by \answerYes{}, \answerNo{}, or \answerNA{}.
    \item[] Justification: {\color{gray} We only utilized public datasets and follow their Terms of use. No new data was introduced in our study. Our model does not have a high risk of being misused.}

\item {\bf Licenses for existing assets}
    \item[] Question: Are the creators or original owners of assets (e.g., code, data, models), used in the paper, properly credited and are the license and terms of use explicitly mentioned and properly respected?
    \item[] Answer: \answerYes{} % Replace by \answerYes{}, \answerNo{}, or \answerNA{}.
    \item[] Justification: {\color{blue} We acknowledged all papers that produced codes and datasets that we used in the paper. The licenses of datasets are provided in the Appendix Section~\ref{Supp:Datasets}}.

\item {\bf New Assets}
    \item[] Question: Are new assets introduced in the paper well documented and is the documentation provided alongside the assets?
    \item[] Answer: \answerNA{} % Replace by \answerYes{}, \answerNo{}, or \answerNA{}.
    \item[] Justification: {\color{gray} Our paper does not release new assets.}

\item {\bf Crowdsourcing and Research with Human Subjects}
    \item[] Question: For crowdsourcing experiments and research with human subjects, does the paper include the full text of instructions given to participants and screenshots, if applicable, as well as details about compensation (if any)? 
    \item[] Answer: \answerNA{} % Replace by \answerYes{}, \answerNo{}, or \answerNA{}.
    \item[] Justification: {\color{gray} Our paper does not involve crowdsourcing nor research with human subjects.}

\item {\bf Institutional Review Board (IRB) Approvals or Equivalent for Research with Human Subjects}
    \item[] Question: Does the paper describe potential risks incurred by study participants, whether such risks were disclosed to the subjects, and whether Institutional Review Board (IRB) approvals (or an equivalent approval/review based on the requirements of your country or institution) were obtained?
    \item[] Answer: \answerNA{} % Replace by \answerYes{}, \answerNo{}, or \answerNA{}.
    \item[] Justification: {\color{gray} Our study does not involve the collection of new datasets containing human subjects.}

\end{enumerate}

\end{document}